\documentclass[acmlarge]{acmart}

\AtBeginDocument{%
	\providecommand\BibTeX{{%
			\normalfont B\kern-0.5em{\scshape i\kern-0.25em b}\kern-0.8em\TeX}}}

\usepackage{booktabs} % For formal tables
\usepackage{multirow}
\usepackage{graphicx}
\usepackage{color}
\usepackage{footnote}
\usepackage{amsmath}  
\usepackage{array}
\usepackage{inputenc}
\usepackage{enumerate}
\usepackage{balance}
\usepackage{amsmath}
\usepackage{amsfonts}
\usepackage{subfigure}
\usepackage{enumitem}
\usepackage{ragged2e}
\usepackage{longtable}
\usepackage{algorithm}
\usepackage{algpseudocode}
\usepackage{multirow}

\usepackage{yfonts}
\usepackage{eufrak}
\usepackage{bbold}

\usepackage{tikz}
\usetikzlibrary{decorations.pathreplacing}
\usepackage{hyperref}

\setcopyright{acmcopyright}
\copyrightyear{2020}
\acmYear{2020}
\acmDOI{10.1145/xxxx.xxxx}

\acmJournal{TKDD}
\acmVolume{xx}
\acmNumber{x}
\acmArticle{xxx}
\acmMonth{9}

\begin{document}

%
% The "title" command has an optional parameter, allowing the author to define a "short title" to be used in page headers.
\title{An Exponential Factorization Machine with Percentage Error Minimization to Retail Sales Forecasting} %for New Products}  %with Long Lead Time and Short Product Life Cycle}

%
% The "author" command and its associated commands are used to define the authors and their affiliations.
% Of note is the shared affiliation of the first two authors, and the "authornote" and "authornotemark" commands
% used to denote shared contribution to the research.
\author{Chongshou Li}
\authornote{corresponding author: Chongshou Li (iselc@nus.edu.sg)}
\email{iselc@nus.edu.sg}
\affiliation{
	\institution{National University of Singapore}
	\streetaddress{1 Engineering Drive 2}
	\city{Blk E1A \#06-25}
	\country{Singapore}
}

\author{Brenda Cheang}
\email{brendacheang@yahoo.com}
\affiliation{
	\institution{Red Jasper Holdings}
	 \streetaddress{Westech Building, 237 Pandan Loop} 
%	\streetaddress{1 Engineering Drive 2}
%	\city{Blk E1A \#06-25}
	\country{Singapore}
}

\author{Zhixing Luo}
\email{luozx.hkphd@gmail.com}
\affiliation{
	\institution{Nanjing University}
	\streetaddress{22 Hankou Road, Gulou District}
	\city{Nanjing}
	\state{Jiangsu}
	\country{China}
}

\author{Andrew Lim}
\email{isealim@nus.edu.sg}
\affiliation{
	\institution{National University of Singapore}
	\streetaddress{1 Engineering Drive 2}
	\city{Blk E1A \#06-25}
	\country{Singapore}
}

%
% By default, the full list of authors will be used in the page headers. Often, this list is too long, and will overlap
% other information printed in the page headers. This command allows the author to define a more concise list
% of authors' names for this purpose.
\renewcommand{\shortauthors}{Li et al.}
%
% The abstract is a short summary of the work to be presented in the article.
\begin{abstract}
This paper proposes a new approach to sales forecasting for new products (stock-keeping units, SKUs) with long lead time but short product life cycle. These SKUs are usually sold for one season only, without any replenishments.  An exponential factorization machine (EFM) sales forecast model is developed to solve this problem which not only takes into account SKU attributes, but also pairwise interactions.  The EFM model is significantly different from the original Factorization Machines (FM) from two-fold: (1) the attribute-level  formulation  for explanatory/input variables  and (2) exponential formulation for the positive response/output/target variable.   The attribute-level formation  excludes  infeasible intra-attribute interactions  and results  in  more efficient feature  engineering comparing with the conventional  one-hot  encoding, while the exponential formulation is demonstrated more effective than the log-transformation for the positive but not  skewed distributed responses.   In order to estimate the parameters, percentage error squares (PES) and error squares (ES) are minimized by a proposed adaptive batch gradient descent (ABGD) method over the training set.   To overcome the over-fitting problem, a greedy forward stepwise feature selection (GFSFS) method is proposed to select the most useful attributes and interactions.   Real-world data provided by a footwear retailer in Singapore is used for testing the proposed approach.   The  forecasting performance in terms  of both  mean absolute percentage  error (MAPE) and mean absolute error (MAE) compares favorably with not only off-the-shelf models but also results reported by  extant sales and demand forecasting studies. The effectiveness of the proposed approach is also demonstrated by  two external public datasets.   Moreover, we prove the theoretical relationships between PES and ES minimization, and present an important property of the PES minimization for regression models; that it trains models to underestimate data. This property fits the situation of sales forecasting where unit-holding cost is much greater than the unit-shortage cost (e.g. perishable products).  
\end{abstract}


%
% Keywords. The author(s) should pick words that accurately describe the work being
% presented. Separate the keywords with commas.

%
% A "teaser" image appears between the author and affiliation information and the body 
% of the document, and typically spans the page. 
%%  \includegraphics[width=\textwidth]{sampleteaser}
%%  \caption{Seattle Mariners at Spring Training, 2010.}
%%  \Description{Enjoying the baseball game from the third-base seats. Ichiro Suzuki preparing to bat.}
% This command processes the author and affiliation and title information and builds
% the first part of the formatted document.
\maketitle

\section{Introduction} \label{intro}
In view of its tremendous impacts to the corporate bottom line, sales forecasting has become less about guessing and more about circumspective calculation. Today, there are a host of predictive methods available to sales forecasting professionals \cite{fader2005value, sanders1994forecasting, madsen2007time}.  However, these methods are typically proposed and implemented in areas where significant relevant demand history is available (examples include assortment planning  \cite{fisher2014demand}, marketing \citep{fader1996modeling}, revenue management \cite{ferreira2015analytics}, and innovation diffusion process \citep{chung2012sales}).  In this vein,  the challenge concerns the development of dependable sales forecasting models for activities that do not have useful historical data to extrapolate from. Consequently, one of the most challenging forecasting problems retailers face is sales forecasting for new products \citep{ching2010designing}.   

Sales forecasting for new products is intricately tied to a retailer's ongoing concern \citep{thomas1993method}.  It carries a high degree of uncertainty and, therefore, risk for the retailer \citep{mentzer2004sales}.   For one, new products typically require a longer lead time to market than existing products because of demand uncertainty; this is counterintuitive as speed to market is essential for the successful launch of new products \citep{ching2010designing}. On the other hand, retailers have to determine the order quantity well before the products reach the stores. Thus, as a result of uncertain demand and long lead time, new products with short life cycles (i.e. seasonal goods) are mostly sold without any replenishments. Therein, miscalculated forecasts could result in significant inventory excesses or shortages.  

With that said, in the absence of product sales history, several forecasting models have been developed using proxy data such as point-of-sales (POS) transaction data of existing stock keeping units (SKUs) and expert opinions on product attributes.   One of the earliest models used was the utility-based multinomial logit (MNL) model by Fader and Hardie \cite{fader1996modeling}.  Fisher and Vaidyanathan \cite{fisher2014demand} developed the exogenous demand model, while Ferreiar et al. \cite{ferreira2015analytics} employed the regression tree  and Chung et al. \cite{chung2012sales} developed the diffusion process based sales forecast model.  However,  these models have their own limitations and the results produced by these models that rely on proxy data have not been excellent to say the least.  And the performance can be largely improved.     Inspired by the aforementioned complexities and the lack of superior solutions, we collaborated with a multinational footwear retailer in Singapore\footnote{The retailer has requested not to release the exact name, which does not influence current study.}   to tackle sales forecasting for new products with long lead time and short life cycle.    Thus, this paper proposes a new method which we have coined the exponential factorization machines (EFM) model to solve this vexing problem.

\begin{figure}[htbp]
	\begin{center}
			\includegraphics[scale = 0.50]{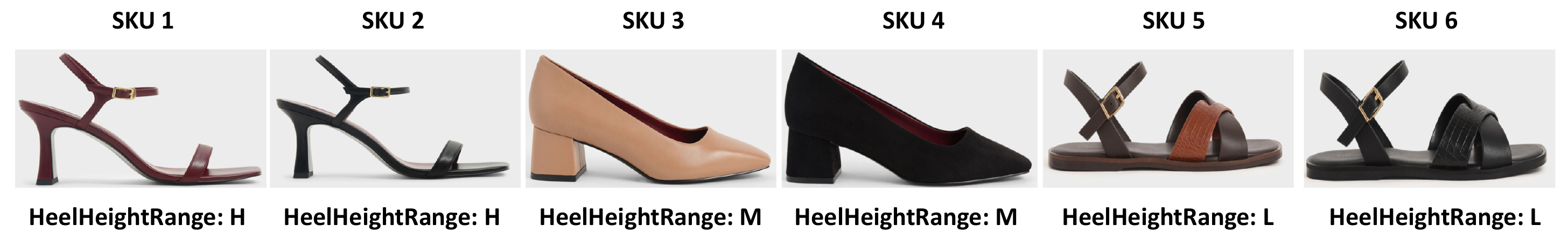}
	\end{center}
	\caption{An illustration of attribute ``\textit{HeelHeightRange}'',  its levels and SKU} \label{Fig:Illustration}
\end{figure}

The products sold by our industrial partner are ladies footwear.   Each product is associated with a set of attributes.  The attributes can be grouped into three categories: (1) visible attributes such as color, size, height etc, (2) latent attributes like designer,  factory (producing the products), (3) marketing behavior such as discount,  average price.  Based on data type, the attributes can also be classified as (1) categorical attributes and (2) numerical attributes. Most attributes are categorical.  For each categorical attribute, there is a fixed set of levels and each product must be associated with one level.  For example, the set of levels for categorical attribute ``\textit{HeelHeightRange}'' is ``\{H, M, L\}'', an illustration is given by Figure \ref{Fig:Illustration}.   In terms of business model,  these ladies footwear can be classified as ``fashion-basic'' products according  to Abernathy  et al. \cite{abernathy1999stitch} and Caro and Martinezde Albeniz \cite{caro2015fast}.   It combines characteristics of  two  classical fast fashion models: (1) fast fashion  and (2) basic products.  The life cycle of these items is as short as eleven weeks sharing the features of the first,  while the lead time is as long as six months which is similar to the second.   After being sold for eleven weeks, the products must be removed from the shelves due.  This makes these fashion products  are ``\textit{perishable}''.  

\begin{figure}[htbp]
	\begin{center}
		\begin{tikzpicture}[scale = 0.65]
		\draw[red, line width = 12] (-5, 0) -- (5, 0);	
		\draw[green, line width = 12] (5, 0) -- (11, 0);
		\draw[thick, <-] (-5, 0.4) -- (-5, 1); 	
		\node[above] at (-5, 1.7) {Forecast total sales over 11 weeks for a new item};
		\node[above] at (-5, 1) {Place the order of intial buy quantity};
		\draw [decoration={brace, mirror, raise = 5pt}, decorate, line width = 1]
		(-5, -0.2) --  (5, -0.2); 
		\node[below] at (0, -0.5) {Lead time of 6 months};
		\draw[thick, <-] (5, 0.4) -- (5, 1); 	
		\node[above] at (5, 1.7) {Release the item};
		\node[above] at (5, 1) {Intital buy on shelves};
		\draw [decoration={brace, mirror, raise = 5pt}, decorate, line width = 1]
		(5, -0.2) --  (11, -0.2); 
		\node[below] at (8, -0.5) {Life time of 11 weeks};
		\draw[thick, <-] (11, 0.4) -- (11, 1); 	
		\node[above] at (11, 1.7) {End of life};
		\node[above] at (11, 1) {Removal from shelves};
		\end{tikzpicture}
	\end{center}
	\caption{Sales forecasting process for a new item of the industry partner} \label{Fig:forecastframework}
\end{figure}
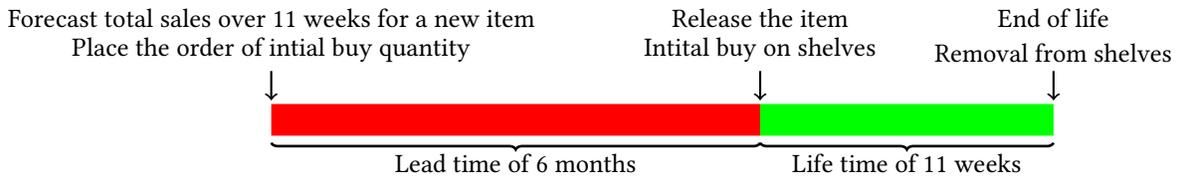

As seen from Figure  \ref{Fig:forecastframework},  we forecast the sales for new products or SKUs (stock keeping units) well before its release. In our study, we forecast sales for the first eleven weeks for each item six months before its launch.  The granularity level of product  we considered is the most fine-grained level, stock keeping unit (SKU).   It is a distinct type of product with unique size and color for sale and Figure \ref{Fig:Illustration} displays six SKUs.   Moreover, we take into account the store locations and  forecast the sales for each new SKU at a store.   The SKU-store sales forecasts are consolidated via the SKU-chain sales which is the sales of each SKU over all stores (the entire chain).  The SKU-chain sales forecasts are used as the new buying quantities.

We approach this conundrum from a machine learning point of view and consider it as a supervised learning problem.  Specifically, the solution is formulated as a regression task where the response (output, target) variable is the sales quantity over 11 weeks for a new SKU at a store (SKU-store sales) and the explanatory variables (factors, features) are product attributes.   Our forecast model is based on the factorization machine (FM)  which is proposed as a generic predictor and have successfully solved problems in many areas such as recommendation systems \citep{rendle2012factorization}.   This model has shown excellent prediction capabilities for categorical data. %Moreover,  it takes interaction between two variables into account and can estimate the effect of new interaction even if the cross is not observed.  
In sales forecasting,  most SKU attributes are categorical.   And interactions between two attributes are found to be important predictors in three industries by Fisher and Vaidyanathan \cite{fisher2014demand}.   Because sales are positive integers in nature,  we generalize the original FM \citep{rendle2012factorization} of order two with an exponential formulation.   In order to estimate parameters of the model,  we propose an adaptive batch gradient descent (ABGD) method to minimizing the gap between forecasts and actual sales over training set.  
The gap is computed by two loss functions: error squares (ES) and percentage error squares (PES). The first is widely used in ordinary least squares (OLS) regression while the second is based on the popular performance indicator, MAPE, in sales and demand forecasting literature.  %An important property of PES is found: it trains model to likely underestimate data comparing with ES.  This property captures practical insights in sales forecasting for perishable products. 
%}

Our contributions are summarized as follows. 

(1) We propose a novel sales forecast model for new SKUs based on historical sales transaction data and attributes data. While previous works have also used historical sales transaction data, none to our knowledge have incorporated SKU attributes and their interactions with the sales transaction data in their approaches.   More importantly, inter-attribute-level interactions which were not observed in the training set can still be estimated in our approach; which differs from polynomial regression with the interaction term.    We incorporated this element in our approach based on the effectiveness observed by Fisher and Vaidyanathan \cite{fisher2014demand}. However, in their work, instead of attribute-level interactions for new SKUs, they constructed the interactions as a new attribute when implementing their model on three industry categories. At the same time, the SKU-store forecast can be used as buying quantities (after consolidation to the SKU-chain) as well as facilitating initial shipment decisions for the store.

(2) We propose an efficient  formulation of the EFM comparing with the formulation of FM \cite{rendle2012factorization}.  %in terms of formulation for the sales forecasting application.   
Firstly,  an important practical  observation from data is that level interactions of the same attribute never appear in both training and test set.   For  example,  attribute ``\textit{size}'' has levels ``$\{S,   M,  L, XL\}$'' and it is impossible to observe the interaction of ``$S$'' and ``$XL$''  on any data  points.   The proposed EFM has removed such infeasible intra-attribute level intersections from the original FM formulation.  It results in more  efficient feature engineering and reduces the search space  comparing than the original FM employing conventional one-hot encoding.    Secondly,  the EFM employs an exponential formulation for the naturally positive response instead of the original FM with log-transformation.  Our computational studies demonstrate that, for positive responses which are not right-skewed distributed,  the exponential formulation is more effective than the log-transformation.  

(3) We study two loss functions of error square (ES) and percentage error square (PES) for parameter estimation.  
Although these two are similar in terms of formulation, an important property of PES is found and demonstrated by the computational study that it likely trains the model to underestimate data.   This property fits the situation where holding cost is much greater than shortage cost.  It is especially true for  perishable products such as fruit, vegetables, fresh meat, milk and fashion products,  the quality of which deteriorate with time and environmental conditions.   For these products,  underestimated demand is better than overestimated demand with the same absolute gap.   Because   overestimation results in buying too much and overstocked items may not be sold before they expire, where the holding costs includes not  only conventional inventory cost but  also  abandonment loss of the expired items. In  this case, PES loss function is a good fit for  demand/sales forecasting.     Moreover, theoretical bounds between objective values of these two optimal solutions are provided (see Theorem 1).    We also introduce a new data normalization technique,  ``instance/sample/row normalization",    which is  different from common normalization methods in data mining \citep{han2011data} and is applicable whenever percentage error minimization is the target under the least square framework for linear models.

(4) We propose a practical and effective feature selection method solution for the desired application.  
Instead of relying on domain knowledge from experts (e.g. working with sales managers in the work of \citep{ferreira2015analytics}), the proposed greedy forward selection method with cross-validation selects a subset of useful attributes and intra-attribute level iterations from all possible candidates, and is based on the proposed exponential factorization machine (EFM) model and historical training data.  
%}

\section{Literature Review} \label{se:litRev}
The proposed exponential factorization machines (EFM) model is based on the factorization machines (FM) model proposed by Rendle \cite{rendle2010factorization}. Likened to support vector machines (SVMs), FMs are said to be excellent general predictors that work for any real-valued feature vector. Nevertheless, FMs are said to be a more superior class of models because they were developed to address the weaknesses of SVMs; by integrating the advantages of SVMs with factorization models \citep{vapnik2013nature}. For instance, unlike SVMs, FMs can model all interactions between variables using factorized parameters (by mapping the interactions to a low dimensional space). Therefore, FMs are capable of modeling $n$-way variable interactions, where $n$ is the number of polynomial order \citep{rendle2012factorization}. However, issues such as numerical stability for some optimization methods have obstructed the generalization of $n$. Thus, the order has been fixed at two as it performs very well in current implementations. As well, FMs not only perform particularly well on sparse data, they are able to compute interactions even in instances with high sparsity \citep{freudenthaler2011bayesian, rendle2012social}. 

In another instance of the superiority of FMs over SVMs, Rendle \cite{rendle2010factorization} established that the model equation of FMs reduce the polynomial computation time to linear complexity, and thus FMs can be optimized directly. At the same time, FMs can mimic specialized factorization models such as parallel factor analysis, matrix factorization, pairwise interaction tensor factorization (PITF) and factorizing personalized Markov chains (FPMC)  that are not applicable as generic predictors as these models are not only derived individually for each task, they require special input data. Thus, the combination of reduced computation time and the ease of application even for users without deep knowledge of factorization models, have made FMs attractive to researchers.

All in all, FMs are popularly applied in collaborative recommendation systems such as recommending movies and music. As they have shown excellent prediction capabilities, they are increasingly being applied to systems such as stock market prediction \citep{chen2014exploiting};  and they are recently extended  with  neural  network as deep  factorization machines for click-through rate (CTR)  prediction and knowledge tracing \citep{guo2017deepfm, vie2018deep,  he2017neural,  xu2019new}.  What motivated us to propose a FM-based approach to our sales forecasting problem are the common elements observed between our problem and collaborative recommendation systems. Firstly, FMs have shown excellent prediction capabilities for categorical data in collaborative recommendation systems. In sales forecasting, most SKU attributes are categorical too. Second of all, the task from both our problem and recommendation systems are regressions.   And as stated earlier, FMs can model all interactions between variables using factorized parameters. And as Fisher   and  Vaidyanathan \citep{fisher2014demand} found this to be important predictors, we believe it is useful in our case as the model would take interaction between two variables into account and can estimate the effects of new interactions even if the inter-attribute level interactions are not observed.  Thus, because sales are positive integers in nature, we employ the exponential formulation of FM and name it as exponential FM (EFM) to it.   
%}

FMs can be trained by three algorithms, namely stochastic gradient descent (SGD) \citep{rendle2010factorization},  alternating least-square (ALS) \citep{rendle2011fast}, and  Markov Chain Monte Carlo (MCMC) \citep{freudenthaler2011bayesian}.   The performance of SGD depends largely on the learning rate, which is one of the hyper-parameters.  The ALS can be applied only when there exists an analytical solution to the minimization problem of parameter estimation. At the same time, both SGD and ALS need to determine regularization hyper-parameters.  At the same time, both SGD and ALS need to determine regularization hyper-parameters.  Rendle et al.  \cite{rendle2012factorization} pointed out that not only does MCMC result in fewer hyper-parameters, but those that need to be specified are not as effective to their initial values. Thus, while the ALS cannot be used because no analytical solutions are viable for the parameter minimization problem for the proposed EFM (due to the exponential formulation), the MCMC is also unsuitable as the response variable in our study (``sales'') does not follow a Gaussian distribution because it is a positive integer in nature. Other distributions such as Poisson are also not found to be a good fit for existing sales data. As a result, we propose an adaptive batch gradient descent (ABGD) algorithm to train the EFM. At each iteration, all training data are used as a batch because the data size is not large and around 5,000. When close to the optimal, ABGD slows down the learning rate to make sure that it converges to the optimum.  

%\blue{
When training the EFM model, we also need to address the feature selection problem.  Features are SKU attributes and attribute interactions.  We modeled SKU sales as attributes as it is commonly used in consumer choice literature where each SKU is defined by a set of attribute levels \citep{fader1996modeling}. Attributes used in this area are salient or consumer recognizable such as package size, color. However, we extend the scope and additionally take both latent attributes (e.g. designer) and marketing attributes (e.g. price and discount percentage) into account. In total, we identified 45 attributes in our database.    %(see Table \ref{tab:attributesExplanations} in Appendix \ref{APP:detailOfAttributes}).  %(see Table 1 in the Supplementary Document). 
And although increasing the number of attributes likely improves the fitting power for training set, it brought upon the problem of how to select the most useful attributes and interactions. If we consider all of them in the model, it would probably result in over-fitting. This is a typical feature selection problem and is known to be NP-hard \citep{guyon2003introduction}.  In the literature,  Chen et al. \cite{chen2014exploiting} and Chen et al. \cite{chen2013general} addressed the problem of feature selection for FM. However, the structure of EFM is different from FM in that it disables the existing feature selection methods.  As such, we propose a greedy forward stepwise feature selection (GFSFS) algorithm with $k$-fold cross-validation to tackle this challenge. %}%%

Finally, there are many performance indicators in terms of forecasting performance evaluation. The most popular ones are mean absolute error (MAE) (or mean absolute deviation (MAD)), mean squared error (MSE) and mean absolute percentage error (MAPE) \citep{kahn1998benchmarking}.  Among other things, the first two indicators are scale-dependent while the third is a scale-independent measure \citep{hyndman2006another}. A survey showed that MAPE and MAE are the top two measures used in the industry; that on average, MAPE of 77\% is acceptable in all industries while MAPE of 76\% is acceptable for consumer product industries \citep{kahn1998benchmarking}.  Makridakis \cite{makridakis1993accuracy}, as well as,  Goodwin and Lawton \cite{goodwin1999asymmetry} also reported that a property of MAPE is that it treats overestimation differently from underestimation.  Therefore, in this study, we develop a percentage error square (PES) loss function based on MAPE and identify its relationships with the classical least square regression. 

%\blue{
\section{Exponential Factorization Machine (EFM) Sales Forecast Model} 
We model the sales quantity for each stock-keeping unit (SKU) of a store by considering both  attributes and the pairwise interactions. Here, we introduce some notations related to the model.  
\begin{itemize}
	\item $N = \{1, 2, \ldots,  n\}$, index set of all SKUs 
	\item $M = \{1, 2, \ldots,  m\}$, index set of all stores 
	\item $A = \{1, \ldots, a\}$, index set of categorical attributes (e.g, color, size)  
	\item $L^i = \{1,\ldots, l_i\}$, $i \in A$, index set of levels of attribute $i$ ($i \in A$) (e.g, ``red'' of color) 
	%\item $I = \{(i,j)~|~i < j, ~i \in L, ~j \in L, \textrm{ level } i \textrm{ and level } j \textrm{ are associated with different attributes}\}$, index set of all possible  pairwise interactions between levels from different attributes (e.g, (``red", ``large") is interaction between ``red'' of color and ``large'' of size)
	\item $I = \{(i, j)~|~i < j,  i \in A,  j \in A\}$, index set of all possible pairwise interactions between categorical attributes of set $A$ (e.g. (``color", ``size") is interaction between ``color" and ``size")
	\item $d_{is}$, response  variable,  actual sales of SKU $i$ at store $s$ for the first eleven weeks after launching,  positive integer ($d_{is} \in \mathbb{N}_{>0}$),  $i \in N, ~ s \in M$
	\item $x^{is}_{cj}$, binary explanatory variable, 1 if SKU $i$ at store $s$ is associated with level $j$ of categorical attribute $c$, and 0 otherwise,  $i \in N$, $s \in M$, $c \in A$ and $j \in L^c$	
	\item $\hat{d}_{is}$,   sales forecast for SKU $i$ at store $s$ for the first eleven weeks after launching, $i \in N, ~ s \in M$ 
	%\item $\hat{y}_{is}$, log of sales forecast for SKU $i$ at store $s$,  $i \in N, ~ s \in M$,  $\hat{y}_{is} = \log(\hat{d}_{is})$ ($\hat{d}_{is} = e^{\hat{y}_{is}}$) 
\end{itemize}
Note that for each categorical attribute, each SKU is for one and only one level. The missing value is treated as a synthetic level.   %The interaction between two levels of the same attribute level can never be observed.  For instance,  interaction between color levels of ``red" and ``blue'' can not be observed  because color of any SKU cannot be ``red'' and ``blue'' simultaneously.  %the color of a SKU cannot be ``red'' and ``blue''  simultaneously.  Therefore,  set $I$ only contains valid interactions each of which consists of two levels associated with two separate attributes. %between levels of different attributes.  %interactions between two levels of a attribute cannot be observed.  For instance, it is impossible to between ``large'' of size and ``small'' of size cannot be observed.    
The proposed EFM  is formulated as follows.
\begin{equation} \label{model:EFM}
\begin{split}
\hat{d}_{is}  =  &  \exp(\beta_0  + \sum_{c \in A} \sum_{j \in L^c} \beta_{cj} x^{is}_{cj}  \\
&  + \sum_{(c,  c^{\prime}) \in I} \sum_{j \in L^c} \sum_{j^{\prime} \in L^{c^{\prime}}}  x^{is}_{cj} x^{is}_{c^{\prime} j^{\prime}} <\mathbf{\mu_{cj}}, ~\mathbf{\mu_{c^{\prime}  j^{\prime}}}>)
\end{split}
\end{equation} 
The sales forecast is generated by an exponential transformation of a factorization machine because  sales data are positive integers, and we name this model as exponential factorization machine (EFM).   The intersection term $\sum_{(c,  c^{\prime}) \in I} \sum_{j \in L^c} \sum_{j^{\prime} \in L^{c^{\prime}}}  x^{is}_{cj} x^{is}_{c^{\prime} j^{\prime}} <\mathbf{\mu_{cj}}, ~\mathbf{\mu_{c^{\prime}  j^{\prime}}}>)$ only considers feasible intra-attribute level intersections.  It reduces feature  space from  $\mathcal{O} (\mathcal{C}^2_{l_m \times a})$ of FM's one-hot encoding to  $\mathcal{O}(\mathcal{C}^2_{a} \times l^2_{m} )$ of the EFM\footnote{$\mathcal{C}^2_{l_m \times a} > \mathcal{C}^2_{a} \times l^2_{m} $ can be easily proved by expanding them and omitted here.},  where $l_m = \max_{i \in A} \{l^i\}$ is the largest number of attribute levels,   $\mathcal{C}^2_y$ is to choose 2 from $y$ and  $\mathcal{C}^2_y = \frac{y\times (y-1)}{2}$. However, the feasible interactions from these two spaces are the same.  The EFM is much more efficient than the FM with one-hot encoding.     Below are parameters (or coefficients) to be estimated.

\begin{itemize}
	\item $\beta_{0}$: global bias (or intercept)
	\item $\beta_{cj}$: parameter for attribute level $j$ of attribute $c$, $c \in A, j \in L^c$ 
	\item $<\mathbf{\mu_{cj}}, ~\mathbf{\mathbf{\mu_{c^{\prime} j^{\prime}}}}>$:   factorized effect on sales of interaction between level $j$ of attribute $c$ and level $j^{\prime}$ of attribute $c^{\prime}$, $(c, c^{\prime}) \in I$, $j \in L^c$ and $j^{\prime} \in L^{c^{\prime}}$.  Here $<\cdot, \cdot>$ is the dot product of two vectors of size $f$ which is a hyper-parameter and also known as  dimensionality of the factorization  \citep{rendle2012factorization}.  It is defined as,  
	\begin{equation} \label{eq:model-factorizatinoM}
	<\mathbf{\mu_{cj}}, ~\mathbf{\mathbf{\mu_{c^{\prime} j^{\prime}}}}> = \sum^f_{p=1} \mu_{cj, p} \mu_{c^{\prime} j^{\prime}, p}.
	\end{equation}	
\end{itemize}
The set of attributes associated with parameter $\mu_{cj, p}$ is denoted as 
\begin{itemize}
	\item $A^{I} = \{c|\exists c^{\prime} \in A \textrm{ such that either }(c, c^{\prime}) \in I \textrm{ or } (c^{\prime}, c) \in I\}$.
\end{itemize}
Currently, because set $I$ contains all possible interactions between any two attributes $c$ and $c^{\prime}$ in set $A$, set $A^I$ is exactly the same as $A$. However, this might result in over-fitting. Thus, we address this issue in Section \ref{sec:fs}, where this notation will be used.  

Now we introduce new notations for learning algorithm development:    $\Theta = \{\beta_0, \beta_{cj}, \mu_{c^{\prime} j^{\prime}, p} | c \in A,  j \in  L^c,  c^{\prime} \in A^{I}, j^{\prime} \in L^{j^{\prime}},  p \in \{1, 2, \ldots, f\}\}$, is a set of model parameters while $\mathbf{x}^{is}$ is a vector of explanatory variables of size $l$ ($l = \sum^a_{i=1} l_i$) for SKU $i$ at store $s$ and $\mathbf{x}^{is} = (x^{is}_{11}, x^{is}_{12}, \ldots, x^{is}_{1(l_1-1)}, x^{is}_{1l_1}, x^{is}_{21}, x^{is}_{22}, \ldots, x^{is}_{2l_2},$ $ \ldots, x^{is}_{a1}, x^{is}_{a2}, \ldots, x^{is}_{al_a})$.  Then sales forecast of SKU $i$ at store $s$, $\hat{d}_{is}$,  can be considered  as a function of explanatory variables and parameters, $\hat{d}_{is} = f(\mathbf{x}^{is}; \Theta)$.  As identified by Rendle et al. \cite{rendle2011fast}, an important property of the factorization machine model is multilinearity.   The customized results for the EFM are: for any $i \in N$,  $s \in M$ and a single parameter $\theta \in \Theta$,    sales forecast $\hat{d}_{is}$ is re-written as, 
\begin{equation} \label{eq:linearitySales}
\hat{d}_{is} = f(\mathbf{x}^{is}; \Theta) = \exp{(\theta h_{(\theta)}(\mathbf{x}^{is}) + g_{(\theta)} (\mathbf{x}^{is}))},
\end{equation}
for every $\theta \in \Theta$.  
Below are explicit formulations of $h_{(\theta)}(\mathbf{x}^{is})$ and $g_{(\theta)}(\mathbf{x}^{is})$ for different $\theta$.  We can see that and both $h_{(\theta)} (\mathbf{x}^{is})$ and $g_{(\theta)} (\mathbf{x}^{is})$ are independent from $\theta$. 
	\begin{itemize}
		\item Case 1: $\theta = \beta_0$
		\begin{equation}
		\left\{
		\begin{array}{ll}
		h_{(\beta_0)} (\mathbf{x}^{is}) = & 1 \\
		g_{(\beta_0)} (\mathbf{x}^{is}) = & \sum_{c \in A} \sum_{j \in L^j} \beta_{cj} x^{is}_{cj} + \sum_{(c,c^{\prime}) \in I} \sum_{j \in L^c} \sum_{j^{\prime} \in L^{c^{\prime}}} x^{is}_{cj} x^{is}_{c^{\prime} j^{\prime}} \sum^f_{p=1} \mu_{cj,p} \mu_{c^{\prime} j^{\prime}, p}
		%\sum_{(j, j^{\prime}) \in I} x^{is}_{j} x^{is}_{j^{\prime}} \sum^f_{p=1} \mu_{j, p} \mu_{j^{\prime}, p} 
		\end{array}
		\right.
		\label{lin:beta_0}
		\end{equation} 
		\item Case 2: $\theta = \beta_{c^{*} j^{*}}, ~ c^{*} \in A, ~ j^{*} \in L^{c^{*}}$
		\begin{equation}
		\left\{
		\begin{array}{ll}
		h_{(\beta_{c^* j^*})} (\mathbf{x}^{is}) = & x^{is}_{c^*j^*} \\
		g_{(\beta_{c^* j^*})} (\mathbf{x}^{is}) = & \beta_0 + \sum_{c \in A, c \neq c^{*}} \sum_{j \in L^c} \beta_{cj} x^{is}_{cj} + \sum_{j \in L^{c^{*}}, j \neq j^{*}} \beta_{c^{*}j} x^{is}_{c^{*}j} + \\
		&  \sum_{(c,c^{\prime}) \in I} \sum_{j \in L^c} \sum_{j^{\prime} \in L^{c^{\prime}}} x^{is}_{cj} x^{is}_{c^{\prime} j^{\prime}} \sum^f_{p=1} \mu_{cj,p} \mu_{c^{\prime} j^{\prime}, p}
		%\sum_{j \in L, j \neq j^*} \beta_j x^{is}_j + \sum_{(j, j^{\prime}) \in I} x^{is}_j x^{is}_{j^{\prime}} \sum^f_{p=1} \mu_{j,p} \mu_{j^{\prime}, p} 
		\end{array}
		\right. \label{lin:beta_jl}
		\end{equation}
		\item Case 3: $\theta = \mu_{c^* j^*, p^{*}},  ~ c^{*} \in A^I,  j^* \in L^{c^{*}},  ~~ p^* \in \{1, \ldots, f\}$ 
		\begin{equation}
		\left\{
		\begin{array}{ll}
		h_{(\mu_{c^*j^*, p^*})} (\mathbf{x}^{is}) = & x^{is}_{c^*j^*} \sum_{(c, c^{*}) \in I \cup (c^{*}, c) \in I} \sum_{j \in L^c} x^{is}_{cj} \mu_{cj, p^{*}} \\ %\sum_{(j, j^*) \in I \cup (j^*, j) \in I} x^{is}_j \mu_{j^*, p^*} \\
		g_{(\mu_{c^*j^*, p^*})} (\mathbf{x}^{is}) = & \beta_0 + \sum_{c \in A} \sum_{j \in L^c} \beta_{cj} x^{is}_{cj} + \\
		& \sum_{j \in L^{c^*}, j \neq j^{*}} \sum_{(c, c^*) \in I \cup (c^*, c) \in I} \sum_{j^{\prime} \in L^c} x^{is}_{c^*j} x^{is}_{cj^{\prime}} \sum^f_{p=1} \mu_{c^* j, p} \mu_{cj^{\prime}, p} + \\
		& \sum_{(c,c^*) \in I \cup (c^*, c) \in I} \sum_{j \in L^c} x^{is}_{c^*j^*} x^{is}_{cj} \sum^f_{p=1, p \neq p^*} \mu_{c^*j^*, p} \mu_{cj,p} + \\
		& \sum_{(c,c^{\prime}) \in I, c\neq c^*, c^{\prime} \neq c^*} \sum_{j \in L^c} \sum_{j^{\prime} \in L^{c^{\prime}}} x^{is}_{cj} x^{is}_{c^{\prime} j^{\prime}} \sum^f_{p=1}  \mu_{cj,p} \mu_{c^{\prime}j^{\prime}, p}
		\end{array}
		\right.
		\label{lin:beta_mu}
		\end{equation}
	\end{itemize}

Now we formalize the data.  % of the training and the holdout test. 
The database is denoted  as $\mathcal{D} = \{(\mathbf{x}^{is}, d_{is})|(i, s)\in S\}$; $\mathbf{x}^{is}$ is a vector of all binary explanatory variables for SKU $i$ at store $s$; $d_{is}$ represents actual sales and also the target (or response) variable;  the observations (or samples,  instances) index set is noted as $S = \{(i, s) | i \in N, s \in M, d_{is} > 0\}$.  Here we only consider valid samples associated with positive sales.  In this study, if a SKU $i$ is not carried at store $s$ ($d_{is} = 0$),  we cannot observe any sales transactions and those observations would not be included in $\mathcal{D}$.  Observations with stock-outs are not considered as well. The rationale is two-fold: 
\begin{enumerate}
	\item One assumption of the EFM forecast model is that items are always available on shelves for the first eleven weeks after a launch.  Including observations with stock-outs violates this assumption.  
	\item Because the time period we considered is the first eleven weeks after launching,  we found that few SKUs were stocked-out during that period in the database.  Instead, the retailer is usually faced with over-stocked quantities in the end of this period and sells them off through markdowns.  Therefore, in our application, there is no need to take stock-outs into account.  
\end{enumerate}
In order to estimate parameters, select features and evaluate model performance, we divide the data into training and holdout test sets which are denoted as $\mathcal{D}^T = \{(\mathbf{x}^{is},d_{is}) | (i, s) \in T \subset S\}$ and $\mathcal{D}^E = \{(\mathbf{x}^{is}, d_{is}) | (i, s) \in E \subset S\}$, respectively.

\section{Estimating the Model Parameters: the Learning Problem} \label{subsec:lp}
Model parameters are estimated by minimizing the gap between forecasts and actual sales over training observations. The gap is computed by \textit{loss} functions.  In order to avoid over-fitting, we apply L2 norm regularization. Resultantly, the model parameters estimation problem becomes an unconstrained minimization problem.  Then, a batch gradient descent algorithm is developed to solve it.

\subsection{Loss Functions}  \label{se:lp:lf}
Two loss functions are  used:  error squares (ES) and  percentage error squares (PES).  Given training set $\mathcal{D}^T = \{(\mathbf{x}^{is}, d_{is})|(i,s) \in T\}$ and corresponding sales forecasts $\{\hat{d}_{is}|(i, s) \in T\}$,  they are defined as follows. 

\begin{itemize}
	\item Error squares (ES):  
	\begin{equation} \label{eq:error:es}
	\mathcal{L}^{\textrm{ES}} (\Theta) = \frac{1}{2} \sum_{(i, s) \in T} (\hat{d}_{is} - d_{is})^2 = \frac{1}{2} \sum_{(i, s) \in T} [f(\mathbf{x}^{is}; \Theta) - d_{is}]^2
	\end{equation}
	\item Percentage error squares (PES): 
	\begin{equation} \label{eq:error:pes}  
  \mathcal{L}^{\textrm{PES}}   (\Theta) = \frac{1}{2} \sum_{(i, s) \in T} (\frac{\hat{d}_{is} - d_{is}}{d_{is}})^2 = \frac{1}{2} \sum_{(i, s) \in T} [\frac{f(\mathbf{x}^{is}; \Theta) - d_{is}}{d_{is}}]^2
	\end{equation}
\end{itemize}

The PES function poses an important property.  Given the same absolute error, the PES loss function penalizes data points with small actual sales more heavily than that with large actual sales.  For instance,  when  actual sales is 120 and forecast is 100 which is underestimated and with an absolute error of 20, the PES function value is $(\frac{20}{120})^2 = (16.67\%)^2 = 0.0278$.  However, when actual sales is 100 and forecast is 120 (which is overestimated with the same absolute error of 20),  the PES function value is  $(\frac{20}{100})^2 = (20\%)^2 = 0.04$ which is greater than $0.0278$. Therefore, in the fitting process of the PES loss function,  data points with small actual values dominate those with large values.  In this light, the PES likely trains the model to underestimate the data, which is demonstrated by our computational results in Section \ref{sec:cor}.  This intuition can also be integrated into the classic newsvendor problem, where the overall loss is usually measured as: 
\begin{equation} \label{eq:news_cost}
\textrm{overall loss} = \textrm{unit holding cost} \times (\widehat{d}_{is} - d_{is})^{+} + \textrm{unit shortage cost} \times (d_{is} - \widehat{d}_{is})^{+}.
\end{equation} 
Here $(\hat{d}_{is} - d_{is})^{+} = \max \{(\hat{d}_{is} - d_{is}), 0\}$. 
The PES loss function can be used when the unit holding cost is much greater than the unit shortage cost (e.g. perishable products).  Different from the above newsvendor loss function which is non-differentiable at $\hat{d}_{is} = d_{is}$ and not applied in current study, the PES is differentiable with respect to the forecast and eases the solution development. 

Now we investigate the relationship between optimalities for these two losses. The following Theorem \ref{theorem:es_pes} demonstrates that it is impossible to minimize the PES and ES loss simultaneously for any training data.  
\begin{theorem} \label{theorem:es_pes}
	Given training set $\mathcal{D}^T = \{(\mathbf{x}_{is}, d_{is}) | (i, s) \in T \}$, forecasts $\{\hat{d}_{is} | (i, s) \in T\}$,  ES minimizer 
	\begin{equation} \label{eq:es:miner}
	\Theta^{*}_{\textrm{ES}} = \operatorname*{arg\, min}_{\Theta \in \mathbb{R}^K}  \mathcal{L}^{\textrm{ES}}( \Theta) = \operatorname*{arg\, min}_{\Theta \in \mathbb{R}^K}  \sum_{(i, s) \in T}{[\hat{d}_{is} - d_{is}]^2}
	\end{equation}
	and PES minimizer 
	\begin{equation} \label{eq:pes:miner}
	\Theta^{*}_{\textrm{PES}} = \operatorname*{arg\, min}_{\Theta \in \mathbb{R}^K}  \mathcal{L}^{\textrm{PES}}( \Theta) = \operatorname*{arg\, min}_{\Theta \in \mathbb{R}^K}  \sum_{(i, s) \in T} {[\frac{\hat{d}_{is} - d_{is}}{d_{is}}]^2}, 
	\end{equation}
	the following results hold:
	
	(a) $\mathcal{L}^{\textrm{ES}}(\Theta^{*}_{PES})$ and $\mathcal{L}^{\textrm{ES}}(\Theta^{*}_{ES})$ satisfy:
	\begin{equation} \label{eq:rel:es:es}
	\mathcal{L}^{\textrm{ES}}(\Theta^{*}_{ES}) \leq \mathcal{L}^{\textrm{ES}}(\Theta^{*}_{PES}) \leq \frac{d^2_{\max}}{d^2_{\min}} ~ \mathcal{L}^{\textrm{ES}}(\Theta^{*}_{ES}),
	\end{equation}
	
	(b) $\mathcal{L}^{\textrm{PES}}(\Theta^{*}_{PES})$ and $\mathcal{L}^{\textrm{PES}}(\Theta^{*}_{ES}) $ satisfy:
	\begin{equation} \label{eq:rel:es:pes}
	\mathcal{L}^{\textrm{PES}}(\Theta^{*}_{PES}) \leq \mathcal{L}^{\textrm{PES}}(\Theta^{*}_{ES}) \leq \frac{d^2_{\max}}{d^2_{\min}} ~ \mathcal{L}^{\textrm{PES}}(\Theta^{*}_{PES}).
	\end{equation}
	Here $\Theta = \{\beta_0, \beta_{cj}, \mu_{c^{\prime} j^{\prime}, p} | c \in A,  j \in L^c, c^{\prime} \in A^I,  j^{\prime} \in L^{c^{\prime}}, p \in \{1, 2, \ldots, f\} \}$, $K = 1 + \sum_{i \in A} l_i + \sum_{i \in A} l_i \times f$, $l_i$ is the number of attribute levels of attribute $i$ ($i \in A$)  and $d_{\max}$ is the maximum actual sales in training set  $\mathcal{D}^T$, $d_{\max} = \max \{d_{is} | (i, s) \in T \}$ while $d_{\min}$ is the minimum,  $d_{\min} = \min \{d_{is} | (i, s) \in T \}$.
\end{theorem}

%See Section B of the Supplementary Document for the proof.  
See Appendix \ref{APP:proof:theorem:es:ps} for the proof.  An interesting observation is the  symmetry structure in above bounds. Minimizing ES %over the training set 
would lead to the non-minimum  PES and the objective value is upper bounded by $\frac{d^2_{\textrm{max}}}{d^2_{\textrm{min}}}$ times of the optimal.  So does minimizing the percentage error squares.  Although above bounds are quite loose,  they are general results for any data $\mathcal{D}^T$ and forecasts $\{\hat{d}_{is} |(i, s) \in T\}$ where no distribution assumptions are  required by 00the data.  Moreover, in the proof,  we do not use any properties of EFM model and learning algorithms, and the results can be extended to other forecast models and learning algorithms. %are applicable for any forecast models and algorithms. %We do not pose any assumptions on them.  %Forecasts can be produced by any models. %We note that the above results are not only for the proposed GFM forecast model but also for any other forecast model.  
With Theorem \ref{theorem:es_pes}, the difference between PES and ES minimization is determined by ratio $\frac{d^2_{\textrm{max}}}{d^2_{\textrm{min}}}$ which can be viewed as an indicator of  data variance.  Particularly, it indicates the distance between the minimal and maximal boundaries of responses, $\{d_{is} | (i, s) \in T\}$, of data $\mathcal{D}^T$.  %which measures the range of data $\mathcal{D}^T$.  
For a trivial case,  if $d_{\textrm{max}} = d_{\textrm{min}}$, all responses in training set are the same.  And the ES minimizer, $\Theta^{*}_{\textrm{ES}}$, is the same as the PES minimizer, $\Theta^{*}_{\textrm{PES}}$.  If $\frac{d^2_{\textrm{max}}}{d^2_{\textrm{min}}}$ is large,  the ES minimizer can be significantly different from the PES minimizer.  We illustrate the theoretical relationships between PES and ES estimation on multiple models and datasets in  Section \ref{sec:cor}. %{\color{blue} We demonstrate this theoretical results of difference between PES and ES estimation for a linear regression model on synthetic data sets in Section \ref{sec:com：ESPES：linear}}.  %}%Simulation investigation for a linear regression model on this issue is provided in Section \ref{sec:com：ESPES：linear}. 

\subsection{Optimization Tasks} \label{sec:ot}
We apply L2 regularization and the parameter estimation becomes the following optimization problem.  
\begin{equation} \label{eq:overallproblem}
\Theta^{*} = \textrm{argmin}_{\Theta \in \mathbb{R}^K}~~ J(\Theta) =  \textrm{argmin}_{\Theta \in \mathbb{R}^K}~~  \mathcal{L}(\Theta) + \frac{1}{2} \sum_{\theta \in \Theta} \lambda_{\theta} \theta^2 
\end{equation} 
Here %the factor $\frac{1}{2}$ is multiplied for later convenience and 
$l(\Theta)$ is  either the ES or the PES loss function.  Below are formulations for the two cases.  
\begin{itemize}
	\item Case 1: optimization task of the ES loss function:
	\begin{equation} \label{eq:opt:es}
	\Theta^{*}_{\textrm{ES}}  = \textrm{argmin}_{\Theta \in \mathbb{R}^K} ~~ J^{\textrm{ES}} (\Theta)
	% =   \mathcal{L}^{\textrm{ES}} (\Theta) 
    % + \frac{1}{2} \sum_{\theta \in \Theta} \lambda_{\theta} \theta^2 
	 =  \textrm{argmin}_{\Theta \in \mathbb{R}^K} ~~ \frac{1}{2} \sum_{(i, s) \in T} (\hat{d}_{is} - d_{is})^2 + \frac{1}{2} \sum_{\theta \in \Theta} \lambda_{\theta} \theta^2. 
	\end{equation}
	\item Case 2: optimization task of the PES loss function:
	\begin{equation} \label{eq:opt:pes}
  \Theta^{*}_{\textrm{PES}}	 = \textrm{argmin}_{\Theta \in \mathbb{R}^{K}} ~~ 
  	J^{\textrm{PES}} (\Theta) % = \mathcal{L}^{\textrm{PES}} (\Theta) } 
    %+ \frac{1}{2} \sum_{\theta \in \Theta} \lambda_{\theta} \theta^2 
    = \textrm{argmin}_{\Theta \in \mathbb{R}^{K}} ~~  \frac{1}{2} \sum_{(i, s) \in T} (\frac{\hat{d}_{is} - d_{is}}{d_{is}})^2 + \frac{1}{2} \sum_{\theta \in \Theta} \lambda_{\theta} \theta^2.
	\end{equation}
\end{itemize}
Here $\lambda_{\theta} \in \mathbb{R}^{+}$  is the regularization hyper-parameter for $\theta$ which controls the regularization importance compared with the loss term.  

\subsection{Adaptive Batch Gradient Descent}
In this subsection, we describe the adaptive batch gradient descent (ABGD) algorithm to solving optimization tasks. The ABGD computes the optimal parameters iteratively. %At each iteration, it looks at all training samples and performs updates simultaneously.  
At each iteration, it looks at all training samples and performs updates simultaneously. For $\theta \in \Theta$,  the rule for parameter update is formulated as, 
\begin{equation}
\theta^{\textrm{new}} \leftarrow \theta^{\textrm{old}} - \eta [\frac{\partial l(\Theta)}{\partial \theta} |_{\theta = \theta^{\textrm{old}}} + \lambda_{\theta}  \theta^{\textrm{old}}],
\end{equation}
where $\eta \in \mathbb{R}^{+}$ is the learning rate (or step size) for gradient descent method while $l(\Theta)$ is a general loss function and either the ES or the PES loss.  Below are gradients for the two cases.%}. 
\begin{itemize}
	\item Case 1: ES loss function
	\begin{equation} \label{eq:descent:es}
	\frac{\partial l^{\textrm{ES}} (\Theta)}{\partial \theta} =  \sum_{(i, s) \in T} (\hat{d}_{is} - d_{is}) \hat{d}_{is} h_{(\theta)}(\mathbf{x}^{is})
	\end{equation}
	\item Case 2: PES loss function
	\begin{equation}  \label{eq:descent:pes}
	\frac{\partial l^{\textrm{PES}} (\Theta)}{\partial \theta} =  \sum_{(i, s) \in T} \frac{(\hat{d}_{is} - d_{is}) \hat{d}_{is}}{d^2_{is}}  h_{(\theta)}(\mathbf{x}^{is})
	\end{equation}
\end{itemize}
Here $\hat{d}_{is}$ is forecast of training sample $(i, s)$ based on current %settings of 
parameters and it is computed by equation (\ref{model:EFM}).  We also utilize the multilinearity of factorization machine for deriving the above results.% and the explicit formulation of $h_{(\theta)}(\mathbf{x}^{is})$ for each $\theta \in \Theta$ is given in Appendix \ref{APP:explicitFormulation}. 
  An observation is that there is a common part in the above derivative for any $\theta$.  It is $(\hat{d}_{is} - d_{is})\hat{d}_{is}$ and $\frac{(\hat{d}_{is} - d_{is})\hat{d}_{is}}{d^2_{is}}$ for the cases of ES and PES loss function, respectively, and does not vary with $\theta$.  At each iteration, we  pre-compute the common term for each sample $(i, s)$ before updating parameters.  As such, the efficiency is improved.   For sample $(i, s)$, we define $v_{is}$ as follows. 
\begin{equation} \label{eq:com:derivate}
\textrm{v}_{is} = \begin{cases}
\eta (\hat{d}_{is} - d_{is}) \hat{d}_{is} & \textrm{if } l(\cdot) = l^{\textrm{ES}}(\cdot) \\
\eta \frac{(\hat{d}_{is} - d_{is})\hat{d}_{is}}{d^2_{is}} & \textrm{if } l(\cdot) = l^{\textrm{PES}}(\cdot)
\end{cases}
\end{equation}
Note that learning rate is taken as the common term as well.  Then the parameter update rule is expressed as 
\begin{equation}
\theta^{\textrm{new}} \leftarrow \theta^{\textrm{old}} - \sum_{(i, s) \in T} v_{is} h_{(\theta)}(\mathbf{x}^{is}) -  \eta \lambda_{\theta}  \theta^{\textrm{old}}.
\end{equation}

Algorithm \ref{alg:bgd} is pseudo-code of ABGD procedure.  %The ABGD algorithm adopts learning rate when the solution is close to the optimal and training error is increased between two iterations.  
At iteration itr, parameters are updated first and then training error  $\textrm{TE}_{\textrm{itr}}$ is computed for the parameters.  The definition of training error varies with the loss functions; %{\color{blue} 
it is mean absolute error (MAE) for the ES loss function and mean absolute percentage error (MAPE) for the PES loss function. %}. %{\color{blue} [Note: here I just want to say if loss function is ES, the definition of training error  $\textrm{TE}_{\textrm{itr}}$ is MAE; if loss function is PES, the definition of training error is MAPE.  (To Do: remove note before submission)]}  
Below is the mathematical formulation. %At each iteration, training error, TE,  is computed for the estimated parameters.  
\begin{equation} \label{eq:trainingerror}
\textrm{TE}_{\textrm{itr}} = \begin{cases}
\frac{1}{|T|} \sum_{(i,s)\in T} |\hat{d}_{is}(\theta^{\textrm{itr}}) - d_{is}| & \textrm{if } l(\cdot) = l^{\textrm{ES}}(\cdot) \\
\frac{1}{|T|} \sum_{(i,s)\in T} |\frac{\hat{d}_{is}(\theta^{\textrm{itr}})  - d_{is}}{d_{is}}| & \textrm{if } l(\cdot) = l^{\textrm{PES}}(\cdot)
\end{cases}
\end{equation}
The ABGD adopts learning rate $\eta$ if two conditions are satisfied: (1) updated solution is close to the optimal, $\textrm{TE}_{\textrm{itr}+1} < \epsilon$,  and (2) training error is increased, $\textrm{TE}_{\textrm{itr}+1} > \textrm{TE}_{\textrm{itr}}$. Note that the threshold value $\epsilon$ is 0.1 for PES loss function while it is 1.0 for ES loss function.  The first condition is necessary because %{\color{blue} 
the gradient descent jumps and can produce solutions with increased training error at the initial iterations. %}. 
The second condition is to make sure that current learning rate is too large to reduce the training error.  %We now 
Note that the stopping criterion in the algorithm is set by the maximum number of iterations, maxInteractions, which is a hyper-parameter  and are generated by the grid search. 

The ABGD solves both $J^{\textrm{ES}}(\Theta)$ and $J^{\textrm{PES}}(\Theta)$ minimization depending on the input loss function type.  If the input loss function is the ES loss $l^{\textrm{ES}}(\Theta)$, it produces the optimal parameters to $J^{\textrm{ES}}$ minimization;  if it is the PES loss $l^{\textrm{PES}}(\Theta)$, it generates the optimal parameters to $J^{\textrm{PES}}$ minimization.  Finally, note that % 
L2 regularization is applied to prevent outlier values of paramters. %values 
% and does a kind of continuous feature selection. %In order to prevent over-fitting, we apply L2 regularization and it does a kind of continuous feature selection.  
However, this does not sufficiently overcome the problem. As a result, we propose a forward stepwise selection algorithm with greedy heuristics and $k$-fold cross validation to select useful features. %which performs some sort of discrete feature selection. 
This is elaborated in the following section.%} %[Note: I do not know how to further explain these centences. Let reviewers to determine whether they are clear or not..... To Do: remove this note before submission.]
%}%However, it is not sufficient to fully overcome this problem.  We propose a forward stepwise selection algorithm with greedy heuristics and $k$-fold cross validation which does a kind of discrete feature selection in the coming section. 
%-----Up to here and write the attribute level selection problem----- %The attribute levels set $L^c$ and attribute interactions set $I^{c}$ should not be the full set of $L$ and $I$,  respectively, because of over-fitting. We address the problem of how to select the best $L^c$ and $I^{c}$ via $k$-fold cross validation in section \ref{sec:fs}.  %words from textbook on subset selection procedure %Elements of statistical learning: By retaining a subset of the predictors %and discarding the rest, subset selection procedures a model that is interpretable  %and has possibly lower prediction error than the full model. However, because it is % a discrete process - variables are either retained or discarded - it often % exhibits high variance, and so doesn't reduce the prediction error of the full model.  % Shrinkage methods are more continuous, and don't suffer as much from high variability. %Thus the lasso does a kind of continuous subset selection. 
\begin{algorithm}[htbp]  \begin{small}
		\caption{Adaptive Batch Gradient Descent} \label{alg:bgd}
		\begin{algorithmic}[1]
			\item[//Inputs:]
			\item[//training data, $\mathcal{D}^T$; subset of attributes, $\textrm{sA}$ ($\textrm{sA} \subset A$); subset of interactions, $\textrm{sI}$ ($\textrm{sI} \subset I$)]
			\item[//loss function $l(\cdot)$, either ES or PES loss]
			%\item[//hyper-parameter for regularization $\{\lambda_{\theta} | \theta \in \Theta\}$]
			%\item[//hyper-parameter: standard deviation,  $\sigma$, of Gaussian distribution for initializing $\mu_{j,p}$]
			%\item[//hyper-parameter: learning rate $\eta$.]
			\item[//Output: the optimal parameters, $\Theta^{*}$, which minimizes either $J^{\textrm{ES}}$ or $J^{\textrm{PES}}$, depending on the input loss.]
			%\Procedure{BGD}{$\mathcal{D}^T, \textrm{sL}, \textrm{sI}, l(\cdot), \sigma, \eta, \{\lambda_{\theta} | \theta \in \Theta\}$} 
			\Procedure{ABGD}{$\mathcal{D}^T, \textrm{sA}, \textrm{sI}, l(\cdot)$}
			\State construct attribute subset of current interactions,  $\textrm{sAI} \gets \{i|\exists j \in A \textrm{ such that either }  (i, j) \in \textrm{sI} \textrm{ or } (j, i) \in \textrm{sI} \}$ %construct set of attribute levels of interactions $I^c$, 
			%$L^{I_c} \gets \{i|\exists j \in L \textrm{ such that either }  (i, j) \in I^c \textrm{ or } (j, i) \in I^c\}$
			\State initialize a step counter, $\textrm{itr} \gets 0$
			\State initialize parameters, $\beta^{\textrm{itr}}_0 \gets 0, ~\beta^{\textrm{itr}}_{ij} \gets 0 ~ \forall i \in \textrm{sA}, ~ j \in L^i, ~\mu^{\textrm{itr}}_{ij,p} \sim \mathcal{N}(0, \sigma) ~ \forall i\in\textrm{sAI}, ~j \in L^i, ~ p\in\{1, 2, \ldots, f\}$
			\State initialize training error of iteration 0, $\textrm{TE}_{0}$, to large number,  $\textrm{TE}_{0} \gets \textrm{Large Number}$
			%\While{stopping criterion is not satisfied (time limit is not exceeded)}
			\While{stopping criterion is not satisfied (iter $\leq$ maxInteractions)} %\item[//Update the global bias, $\beta_0$]
			\State for each $(i, s) \in T$, compute $v_{is}$ by equation (\ref{eq:com:derivate}) 
			\State $\beta_0^{\textrm{itr} + 1} \gets \beta_0^{\textrm{itr}} -   \sum_{(i,s) \in T} v_{is} h_{(\beta_0)}(\mathbf{x}^{is})  -  \eta \lambda_{\beta_0} \beta^{\textrm{itr}}_0$
			%\item[//Update $\beta_{jl}$]
			\For{$c \in \textrm{sA}, j \in L^i$}
			\State $\beta^{\textrm{itr} + 1}_{cj} \gets \beta^{\textrm{itr}}_{cj} -  \sum_{(i,s) \in T} v_{is} h_{(\beta_{cj})}(\mathbf{x}^{is})  -   \eta \lambda_{\beta_{cj}} \beta^{\textrm{itr}}_{cj}$
			\EndFor
			\For{$c \in \textrm{sAI}, j \in L^c,  p \in \{1, 2, \ldots, f\}$}
			\State $\mu^{\textrm{itr} + 1}_{cj,p} \gets  \mu^{\textrm{itr}}_{cj,p}  -  \sum_{(i,s) \in T} v_{is} h_{(\mu_{cj, p})}(\mathbf{x}^{is})   - \eta \lambda_{\mu_{cj, p}} \mu^{\textrm{itr}}_{cj, p}$
			\EndFor
			\State compute training error, $\textrm{TE}_{\textrm{itr}+1}$, based on parameters $\{\theta^{\textrm{itr}+1}\}$ following equation (\ref{eq:trainingerror})
			\If{$\textrm{TE}_{\textrm{itr}+1} < \epsilon $ (solution is close to the optimal) and $\textrm{TE}_{\textrm{itr}+1} > \textrm{TE}_{\textrm{itr}}$}
			%\If{$\{\textrm{cvJ}^{\textrm{itr}}_i | i \in \{1, 2, \ldots, k\}\}$ is significantly smaller than $\{\textrm{cvJ}^*_i ~|~ i \in \{1, 2, \ldots, k\}\}$}
			\State $\eta \gets \eta/2$  %$\textrm{sA}^* \gets \textrm{currentA}^{\textrm{itr}}$,  $\textrm{sI}^* \gets \textrm{currentI}^{\textrm{itr}}$, $\textrm{cvJ}^*_i \gets \textrm{cvJ}^{\textrm{itr}}_i$ $\forall i \in \{1, 2, \ldots, k\}$ \label{gfsfs:update}
			\EndIf
			
			\State $\textrm{TE}_{\textrm{itr}} \gets \textrm{TE}_{\textrm{itr+1}}$,  $\textrm{itr} \gets \textrm{itr} + 1$
			\EndWhile
			\State \textbf{Return} $\Theta^{*} = \{\beta^{\textrm{itr}}_0, \beta^{\textrm{itr}}_{cj}, \mu^{\textrm{itr}}_{c^{\prime} j^{\prime},p} | c \in \textrm{sA}, j \in L^c, c^{\prime} \in \textrm{sAI},  j^{\prime} \in L^{c^{\prime}}, p \in \{1, 2, \ldots, f\} \}$
			\EndProcedure
		\end{algorithmic}
	\end{small}
\end{algorithm}

\section{Training and Forecasting} \label{sec:trainingandtest}
\subsection{Feature Selection} \label{sec:fs}
Feature selection is traditionally known as a challenging problem in machine learning.  In the EFM model, there are two kinds of features:  (1) single attributes  $A$ and (2) pairwise interactions between attributes $I$.   In a training set for a class of around 5,000 observations, there are 45 attributes and 990 attribute interactions ($C^2_{45} = \frac{45 \times 44}{2} = 990$).   %{\color{blue} 
Sizes of training and test sets are provided by Table \ref{tab:descriptiveStatistics} of Section \ref{sec:dataIntro}. %} %Exact descriptive statistics are referred to Table \ref{tab:descriptiveStatistics} in Section  \ref{sec:dataIntro}. %(To Do: add Section Number). %(To Do: double-check these number and finally confirm them).  
Putting all features into the model can result in minimal losses over training set but likely produces large forecasting error for the hold-out test set.   Not all of them are useful and only subsets of $A$ and $I$ should be taken into account.   In order to address this over-fitting problem, we propose a greedy forward stepwise feature selection (GFSFS) algorithm.  The GFSFS searches feature subset space iteratively. %step by step.  
At each iteration,  it augments features by greedily adding a subset of either attributes or pairwise interactions which has the greatest \textit{potential} to improve forecasting accuracy over the test set.  %Its detail is given in Algorithm \ref{alg:FS} and subsection \ref{sec:fs:fss} (To Do: double-check and revise this sentence).  
Once a new setting of features is built, we investigate how good it is.  Although the aim is to find the setting which produces smallest forecasting error over the hold-out test set,  the test set is unseen in the training stage. %}. 
Thus, a $k$-fold cross-validation (CV) is employed to estimate the error. And the goodness is quantified by validation errors of the CV.  %which is present by sub-procedure Algorithm \ref{alg:CV} and subsection \ref{sec:fs:cv}.  
%In the end, a subset of levels and a subset of interactions are generated by the GFSFS. 
In the end, a feature setting of either a subset of attributes $A$ or a subset of interactions $I$ is generated by the GFSFS.  %It results in the minimal validation errors among all subsets of features searched.   
This resulting subset of attributes is denoted as $\textrm{sA}^*$  while the subset of interactions is represented by $\textrm{sI}^*$,  and the corresponding minimal validation errors of the CV is noted as $\{\textrm{cvJ}^*_i | i \in \{1, 2, \ldots, k\}\}$. %associated minimal validation errors is stored as output of the GFSFS. And these two subsets are denoted as, $\textrm{sL}^*$ and $\textrm{sI}^*$, respectively.  
The GFSFS is displayed by Algorithm \ref{alg:GFSFS} and its sub-procedures are present in the following subsections.  %,in the GFSFS. 

\begin{algorithm}[htbp]
	\begin{small}
		\caption{Greedy Forward Stepwise Feature Selection} \label{alg:GFSFS}
		\begin{algorithmic}[1]
			\item[//Inputs:]
			\item[//training data, $\mathcal{D}^T$; set of all possible attributes, $A$; set of all possible interactions, $I$]
			\item[//loss function $l(\cdot)$, either ES or PES loss]
			%\item[//number of folds for cross validation: $k$]
			%\item[//hyper-parameters: $\sigma$, $\eta$, $\{\lambda_\theta | \theta \in \Theta\}$]
			\item[//Output: $(\textrm{sA}^*, \textrm{sI}^*)$, the optimal subsets of attributes $A$ and interactions $I$] %(or subset of levels $L$ and interactions $I$)]
			\Procedure{GFSFS}{$\mathcal{D}^T, L, I, l(\cdot)$} %k, \sigma, \eta, \{\lambda_\theta | \theta \in \Theta\}$} 
			\State randomly partition training set $\mathcal{D}^T$ into $k$ complementary groups with roughly equal size,  $\Lambda = \{T_i , i \in \{1, 2, \ldots, k\}| T_1 \cup T_2 \cup T_3 \cup \cdots \cup T_k = T,  T_j \cap T_{j^\prime} = \emptyset, j \neq j^{\prime},  j, j^{\prime} \in \{1, 2, \ldots, k\}\}$\label{gfsfs:partition} 
			\State initialization, $\textrm{currentA}^0 \gets \emptyset$, $\textrm{currentI}^0 \gets \emptyset$, $\textrm{sd}^1 \gets 1$, $\textrm{isFeasible}_{-1} \gets \textrm{True}$, $\textrm{isFeasible}_1 \gets \textrm{True}$,  $\textrm{itr} \gets 1$\label{gfsfs:initialization}
			\State for each $i \in {1, 2, \ldots, k}$, initialize the optimal validation error $\textrm{cvJ}^*_i$ to be $\textrm{cvJ}^{\textrm{null}}_{i}$ given by equation (\ref{eq:optimal:initial})\label{gfsfs:initializeVE} 
			\While{current search direction is feasible ($\textrm{isFeasible}_{\textrm{sd}^\textrm{itr}}$ is True)}\label{gfsfs:condStop}
			%\State feature subset selection, $\{\Delta L, \Delta I\} \gets \textrm{FSS}(\textrm{sd}^{\textrm{itr}}, \textrm{currentL}^{\textrm{itr}-1},  \textrm{currentI}^{\textrm{itr}-1}, L, I)$	
			\State feature subset selection, $\{\Delta A, \Delta I\} \gets \textrm{FSS}(\textrm{sd}^{\textrm{itr}}, \textrm{currentA}^{\textrm{itr}-1},  \textrm{currentI}^{\textrm{itr}-1})$\label{gfsfs:fss}	
			\State $\textrm{currentA}^{\textrm{itr}} \gets \textrm{currentA}^{\textrm{itr} - 1} \cup \Delta A$,  $\textrm{currentI}^{\textrm{itr}} \gets \textrm{currentI}^{\textrm{itr} - 1} \cup \Delta I$\label{gfsfs:constructNew}
			\If{either $\Delta A$ or $\Delta I$ is non-empty}		
			\State perform cross validation,  $\{\textrm{cvJ}^{\textrm{itr}}_{i} | i \in \{1, 2, \ldots, k\}\} \gets$ CV($\textrm{currentA}^{\textrm{itr}}$, $\textrm{currentI}^{\textrm{itr}}$, $k$, $l(\cdot)$, $\Lambda$)\label{gfsfs:subCV} 
			\If{$\{\textrm{cvJ}^{\textrm{itr}}_i | i \in \{1, 2, \ldots, k\}\}$ is significantly smaller than $\{\textrm{cvJ}^*_i ~|~ i \in \{1, 2, \ldots, k\}\}$}
			\State $\textrm{sA}^* \gets \textrm{currentA}^{\textrm{itr}}$,  $\textrm{sI}^* \gets \textrm{currentI}^{\textrm{itr}}$, $\textrm{cvJ}^*_i \gets \textrm{cvJ}^{\textrm{itr}}_i$ $\forall i \in \{1, 2, \ldots, k\}$ \label{gfsfs:update}
			\State $\textrm{isFeasible}_{1} \gets \textrm{True}$, $\textrm{isFeasible}_{-1} \gets \textrm{True}$\label{gfsfs:reinitialize}	%, $\textrm{sd}^{\textrm{itr}+1} \gets \textrm{sd}^{\textrm{itr}}$\label{gfsfs:reinitialize}		
			\Else
			\State $\textrm{isFeasible}_{\textrm{sd}^{\textrm{itr}}} \gets \textrm{False}$,  %, $\textrm{sd}^{\textrm{itr} + 1}  \gets - \textrm{sd}^{\textrm{itr}}$\label{gfsfs:nonoptimal} %\State 
			$\textrm{currentA}^{\textrm{itr}} \gets \textrm{currentA}^{\textrm{itr} - 1}$, $\textrm{currentI}^{\textrm{itr}} \gets \textrm{currentI}^{\textrm{itr} - 1}$\label{gfsfs:nonoptimal}
			\EndIf
			\Else
			\State $\textrm{isFeasible}_{\textrm{sd}^{\textrm{itr}}} \gets \textrm{False}$\label{gfsfs:infeasible}%, $\textrm{sd}^{\textrm{itr} + 1}  \gets  - \textrm{sd}^{\textrm{itr}}$ \label{gfsfs:infeasible}
			\EndIf	
			\If{search direction $- \textrm{sd}^{\textrm{itr}}$ is feasible ($\textrm{isFeasible}_{- \textrm{sd}^\textrm{itr}}$ is True)} \label{gfsfs:dfs:b}
			\State $\textrm{sd}^{\textrm{itr} + 1}  \gets - \textrm{sd}^{\textrm{itr}}$
			\EndIf \label{gfsfs:dfs:e}
			\State $\textrm{itr} \gets \textrm{itr} + 1$
			%Until{No feasible directions found ($\textrm{direction}  == -1$)}
			\EndWhile
			\State \textbf{Return} $(\textrm{sA}^*, \textrm{sI}^*)$
			\EndProcedure
		\end{algorithmic}
	\end{small}
\end{algorithm}

The GFSFS initializes the optimal subsets of attributes, $\textrm{sA}^*$, and interactions, $\textrm{sI}^*$,  to be empty (see line \ref{gfsfs:initialization}). %at Algorithm \ref{alg:GFSFS}).   
The model without any features is named as null model where only global bias (or intercept) is considered.   There exist analytical solutions for this model and corresponding errors of  $k$ validation sets can be directly derived instead of calling sub-procedure CV.  The detail is referred to subsection \ref{sec:fs:null}.   Line \ref{gfsfs:initializeVE} %at the GFSFS
initializes $\{\textrm{cvJ}^*_i |  i \in \{1, 2, \ldots, k\}\}$  to be validation errors of the null model.  After that, the GFSFS constructs new subsets of features iteratively.  The iteration %(or step) 
number is denoted as itr which is initialized to be one. (Null model is investigated at iteration zero.)  At iteration itr, the GFSFS augments current feature subsets of either  attributes, $\textrm{currentA}^{\textrm{iter}-1}$, or interactions, $\textrm{currentI}^{\textrm{itr}-1}$.  There are two choices: %for this operation: 
(1) adding attributes and (2) adding interactions.  These two choices are noted as search directions at iteration itr and symbolized as  $\textrm{sd}^{\textrm{itr}}$,  %$\textrm{sd}^{\textrm{itr}}$ 
which is defined as an enumerated variable in the GFSFS and takes two possible values of positive one and negative one.  Being positive one means adding attributes while taking value of negative one represents adding interactions. A trick of this setting is that search direction can be easily changed via multiplying search direction by negative one.   The search direction for iteration one is initialized to be positive one of adding attributes, $\textrm{sd}^{1} \gets 1$ (see line \ref{gfsfs:initialization}).   At iteration itr,  depending on search direction $\textrm{sd}^{\textrm{itr}}$, the sub-procedure FSS greedily selects a new subset of either attributes $\Delta L$ or interactions $\Delta I$  for augmenting current features at line \ref{gfsfs:fss}.  Subsection \ref{sec:fs:fss} presents this sub-procedure in detail.  

At the same time, there exists a feasibility issue for search directions.  The GFSFS cannot always add either attributes or interactions to current features.  %Not all of these two search directions are always feasible.  
For instance, at iteration itr,  given that search direction, $\textrm{sd}^{\textrm{itr}}$, is positive one of adding attributes and current subset of attributes, $\textrm{currentA}^{\textrm{itr} - 1}$, contains all attributes  ($\textrm{currentA}^{\textrm{itr} - 1} = A$),  there is no attributes which are out of current subset and it is impossible to augment current features following the given direction. And the direction is not feasible.  In order to formalize it, the GFSFS defines boolean variables for labeling the feasibility of two search directions.  They are $\textrm{isFeasible}_{1}$ and $\textrm{isFeasible}_{-1}$ for positive one of adding attributes and negative one of adding interactions, respectively.  A value of True means that it is feasible while a value of False means that it is infeasible.  These two boolean variables are initialized to be True (feasible) at line \ref{gfsfs:initialization}.   

At iteration itr, based on search direction $\textrm{sd}^{\textrm{itr}}$, attribute subset $\textrm{currentA}^{\textrm{itr}-1}$ and interaction subset $\textrm{currentI}^{\textrm{itr}-1}$,  (in line \ref{gfsfs:fss}) the FSS produces  one subset of either attributes, $\Delta A$,  or interactions,  $\Delta I$, only for augmenting  features at iteration $\textrm{itr} - 1$.  Then features for current iteration itr are constructed via taking union of features of previous iteration and newly found increments,  $\textrm{currentA}^{\textrm{itr}} \gets \textrm{currentA}^{\textrm{itr} - 1} \cup \Delta A$,  $\textrm{currentI}^{\textrm{itr}} \gets \textrm{currentI}^{\textrm{itr} - 1} \cup \Delta I$, which is seen in line \ref{gfsfs:constructNew}.   If both $\Delta A$ and $\Delta I$ are empty sets,  it indicates that  current search direction $\textrm{sd}^{\textrm{itr}}$ is not feasible any more,  $\textrm{isFeasible}_{\textrm{sd}^{\textrm{itr}}} \gets \textrm{False}$ (see line \ref{gfsfs:infeasible}). %and search direction should be changed for next iteration, $\textrm{sd}^{\textrm{itr}+1} \gets - \textrm{sd}^{\textrm{itr}}$. It is implemented at line \ref{gfsfs:infeasible}. 
If either $\Delta A$ or $\Delta I$ is not empty,  $k$-fold cross validation is performed for examining the goodness of this new setting at line \ref{gfsfs:subCV}.  If the mean of $k$ validation errors,  $\{\textrm{cvJ}^{\textrm{itr}}_i | i \in \{1, 2, \ldots, k\}\}$,  is smaller than that of the current optimal validation errors, $\{\textrm{cvJ}^{*}_i | i \in \{1, 2, \ldots, k\}\}$,  at a significance level of 0.05 by a paired t-test,  the optimal subsets and validation errors are updated at line \ref{gfsfs:update},  $\textrm{sA}^* \gets \textrm{currentA}^{\textrm{itr}}$, $\textrm{sI}^* \gets \textrm{currentI}^{\textrm{itr}}$,   $\textrm{cvJ}_i^* \gets \textrm{cvJ}^{\textrm{itr}}_i ~\forall i \in \{1, 2, \ldots, k\}$.  Line \ref{gfsfs:reinitialize} sets two search directions to be feasible  
%Besides,  two search directions are set to be feasible,  
$\textrm{isFeasible}_{1} \gets \textrm{True}$  and $\textrm{isFeasible}_{-1} \gets \textrm{True}$. %,and keeps search direction for next iteration unchanged, %$\textrm{sd}^{\textrm{itr} + 1} \gets \textrm{sd}^{\textrm{itr}}$.  
On the other hand, if average of $k$ validation errors is not significantly smaller, current search direction is set to be infeasible, $\textrm{isFeasible}_{\textrm{sd}^{\textrm{itr}}} \gets \textrm{False}$, and feature subsets remain the same, $\textrm{currentA}^{\textrm{itr}} \gets \textrm{currentA}^{\textrm{itr}-1}$ and $\textrm{currentI}^{\textrm{itr}} \gets \textrm{currentI}^{\textrm{itr}-1}$ (see line \ref{gfsfs:nonoptimal}). %,and search for next iteration is changed,  $\textrm{sd}^{\textrm{itr}+1} \gets - \textrm{sd}^{\textrm{str}}$, which is referred to line \ref{gfsfs:nonoptimal}. And settings of feature remain the same (see line \ref{gfsfs:nochangesettings}).    
The GFSFS employs a depth-first search (DFS) strategy.  At each iteration, the search direction is changed from that of last iteration if it is feasible at lines \ref{gfsfs:dfs:b} - \ref{gfsfs:dfs:e}.   The GFSFS stops when current search direction is not feasible (see line \ref{gfsfs:condStop}). 

The $k$-fold cross-validation %for all subsets searches by GFSFS 
uses the same partition $\Lambda$ of training set $\mathcal{D}^T$ which is defined at the beginning of the GFSFS algorithm (line \ref{gfsfs:partition}).  As such,  $k$ validation sets do not vary with iterations and $t$-test is employed for the comparison.

\subsection{Cross-Validation} \label{sec:fs:cv}
In this subsection, we outline the $k$-fold cross-validation (CV) in Algorithm \ref{alg:CV} which is a sub-procedure and used by the GFSFS algorithm.   In $k$-fold cross-validation procedure, the model is fitted and evaluated $k$ times.  At iteration $i$, fold $i$  is held out  and used as validation set noted as $\mathcal{D}^{\textrm{cE}}$.  And the remaining $k-1$ folds are used as training set noted as $\mathcal{D}^{\textrm{cT}}$.   Parameters for selected subset of features are estimated via calling the ABGD procedure.  %We note that parameters are estimated by calling the BGD procedure.   
And the validation error is computed and noted as $\textrm{cvJ}_i$.   The error type depends on the loss function $l(\cdot)$.  If it is the ES loss function, validation error $\textrm{cvJ}_i$ is the mean absolute error (MAE) between forecasts and actual sales over validation set $\mathcal{D}^{\textrm{cE}}$; otherwise, it is the mean absolute percentage error (MAPE).  It is given as follows.  
\begin{equation} \label{eq:cv:error}
\textrm{cvJ}_i = \begin{cases}
\frac{1}{|T_i|} \sum_{(j,s)\in T_i} |\hat{d}_{js} - d_{js}| & \textrm{if } l(\cdot) = l^{\textrm{ES}}(\cdot) \\
\frac{1}{|T_i|} \sum_{(j,s)\in T_i} |\frac{\hat{d}_{js} - d_{js}}{d_{js}}| & \textrm{if } l(\cdot) = l^{\textrm{PES}}(\cdot)
\end{cases}
\end{equation}
%\begin{equation} \label{eq:cv:es:error} %\textrm{cvJ}_i = %\end{equation} %If it is the PES loss function, validation error is the mean absolute percentage error (MAPE) and it is %\begin{equation} \label{eq:cv:pes:error} %\textrm{cvJ}_i =  %\end{equation} 
In the end, we note that, as a rule of thumb, $k$ is set to be five. %and five-fold cross validation  is applied. 
\begin{algorithm}[htbp]
	\begin{small}
		\caption{$k$-fold Cross-Validation} \label{alg:CV}
		\begin{algorithmic}[1]
			\item[//Inputs:]
			\item[//attribute level subset, csL; interaction subset, csI]
			%\item[//number of folds for validation, $k$; 
			\item[//partition of training data, $\Lambda = \{T_i, i \in \{1,\ldots, k\}\}$; loss function $l(\cdot)$]
			%\item[//] 
			\item[//Outputs:]
			\item[//objective values of $k$ validation sets $\{\textrm{cvJ}_i | i \in \{1, 2, \ldots, k\} \}$] 
			\Procedure{CV}{csL, csI, %$k$, 
				$l(\cdot)$, $\Lambda$} %$\sigma$, $\eta$, $\{\lambda_\theta | \theta \in \Theta\}$} 
			\For{$i = 1 \to k$}
			\State $\mathcal{D}^{\textrm{cE}} \gets \{(\mathbf{x}^{js}, d_{js})| (j, s) \in T_i\}$,  $\mathcal{D}^{\textrm{cT}} \gets \emptyset$ 
			\For{$g = 1 \to k$}
			\If{$g \neq i$}
			\State $\mathcal{D}^{\textrm{cT}} \gets \mathcal{D}^{\textrm{cT}} \cup \{(\mathbf{x}^{js}, d_{js})| (j, s) \in T_g\}$
			\EndIf
			\EndFor
			\State estimate parameters by the ABGD, $\Theta^{c} \gets $ ABGD($\mathcal{D}^{\textrm{cE}}$, csL, csI,  $l(\cdot)$) %$\sigma$, $\eta$, $\{\lambda_{\theta}|\theta \in \Theta\}$) 
			\State based on input loss function $\l(\cdot)$, compute validation error, $\textrm{cvJ}_i$, following equation(\ref{eq:cv:error}) %or (\ref{eq:cv:pes:error}) 
			\EndFor
			\State \textbf{Return} $\{\textrm{cvJ}_i | i \in \{1, 2, \ldots, k\}\}$
			\EndProcedure
		\end{algorithmic}
	\end{small}
\end{algorithm}

\subsection{Null Model} \label{sec:fs:null}
The null model is used for initialization in feature selection.  In the null model, no attributes and interactions are taken into account.  Thus, only global bias (or intercept) $\beta_0$ is required to be estimated and all other parameters are set to be zero.  For training sample $(i, s) \in T$, forecast $\hat{d}_{is}$ by null model is: 
%\blue{
\begin{equation*}
\hat{d}_{is} = f^{\textrm{null}}(\mathbf{x}^{is}, \Theta) = \exp({\beta_0}).
\end{equation*}
%}
%\blue{
Given training set $\mathcal{D}^T$, we do not consider regularization for estimating $\beta_0$.  And two optimization tasks are:
\begin{itemize}
	\item Case 1: optimization task of the ES loss for null model
	\begin{equation}  \label{eq:tasks:es:null}
	J^{\textrm{ES}}_{\textrm{null}} (\beta_0) = \frac{1}{2} \sum_{(i, s) \in T} [\exp({\beta_0}) - d_{is}]^2
	\end{equation}
	\item Case 2: optimization task of the PES loss for null model
	\begin{equation} \label{eq:tasks:pes:null}
	J^{\textrm{PES}}_{\textrm{null}} (\beta_0) = \frac{1}{2} \sum_{(i, s) \in T} [ \frac{\exp(\beta_0)}{d_{is}} - 1]^2.
	\end{equation}
\end{itemize}
%}

Taking the first derivate and setting it to be zero, we can get analytic solutions to minimizing these two tasks.  The optimal parameter $\beta^{\textrm{null}*}_{0, \textrm{ES}}$ to $J^{\textrm{ES}}_{\textrm{null}} (\beta_0)$ minimization is 
\begin{equation} \label{eq:analytical:es:solution}
\beta^{\textrm{null}*}_{0,\textrm{ES}} = \operatorname*{arg\, min}_{\beta_0 \in \mathbb{R}}  J^{\textrm{ES}}_{\textrm{null}} (\beta_0) = \log [\frac{\sum_{(i,s)\in T} d_{is}}{|T|}].
\end{equation}
The optimal parameter $\beta^{\textrm{null}*}_{0, \textrm{PES}}$ to  $J^{\textrm{PES}}_{\textrm{null}}(\beta_0)$ minimization is 
\begin{equation} \label{eq:analytical:pes:solution}
\beta^{\textrm{null}*}_{0,\textrm{PES}} = \operatorname*{arg\, min}_{\beta_0 \in \mathbb{R}} J^{\textrm{PES}}_{\textrm{null}} (\beta_0)   =  \log[\sum_{(i,s) \in T}\frac{1}{d_{is}}] - \log[\sum_{(i, s)\in T} \frac{1}{d^2_{is}}].
\end{equation}
Based on these two solutions and partition $\Lambda$ of training set $\mathcal{D}^T$, outputs of the CV can be derived and below are validation error $\textrm{cvJ}^{\textrm{null}}_i$ for data set $T_i$ for two different loss functions. 
\begin{equation} \label{eq:optimal:initial}
\textrm{cvJ}^{\textrm{null}}_i = \begin{cases}
\frac{1}{|T_i|} \sum_{(i, s)\in T_i} |d^{\textrm{ESM}}_{T\backslash T_i} - d_{is}|  & \textrm{ if } l(\cdot) = l^{\textrm{ES}}(\cdot)\\
\frac{1}{|T_i|} \sum_{(i, s) \in T_i} |\frac{d^{\textrm{PESM}}_{T \backslash T_i} - d_{is}}{d_{is}}| & \textrm{ if } l(\cdot) = l^{\textrm{PES}}(\cdot)
\end{cases}
\end{equation}
Here $d^{\textrm{ESM}}_{T \backslash T_i} = \frac{\sum_{(i,s)\in T \backslash T_i} d_{is}}{|T \backslash T_i |}$ and $d^{\textrm{PESM}}_{T \backslash T_i} = \frac{\sum_{(i,s)\in T \backslash  T_i} \frac{1}{d_{is}}}{\sum_{(i, s)\in T \backslash T_i} \frac{1}{d^2_{is}}}$. 
%If loss function is the ES, validation error $\textrm{cvJ}_i$ is  %\begin{equation} \label{eq:es:optimal:initial}  %\textrm{cvJ}_i = \frac{1}{|T_i|} \sum_{(i, s)\in T_i} |d^{\textrm{ESM}}_{T\backslash T_i} - d_{is}| %\end{equation} %where $d^{\textrm{ESM}}_{T \backslash T_i} = \frac{\sum_{(i,s)\in T \backslash T_i} d_{is}}{|T \backslash T_i |}$; if loss function is the PES, validation error $\textrm{cvJ}_i$ is  %\begin{equation} \label{eq:pes:optimal:initial}  %\textrm{cvJ}_i = \frac{1}{|T_i|} \sum_{(i, s) \in T_i} |\frac{d^{\textrm{PESM}}_{T \backslash T_i} - d_{is}}{d_{is}}| \end{equation} %where $d^{\textrm{PESM}}_{T \backslash T_i} = \frac{\sum_{(i,s)\in T \backslash  T_i} \frac{1}{d_{is}}}{\sum_{(i, s)\in T \backslash T_i} \frac{1}{d^2_{is}}}$. The null model is first investigated and above results are used to initialize $k$ validation errors at line \ref{gfsfs:initializeVE} of the GFSFS algorithm. 
%-----------------------Up to Here------------------------

\subsection{Forward Subset Selection} \label{sec:fs:fss}
In this subsection, we present the forward subset selection (FSS) algorithm (see Algorithm \ref{alg:FS}), a key aspect of the GFSFS.  The FSS greedily finds a subset of attributes $\Delta A$ or interactions $\Delta I$ (depending on input search direction sd) in an effort to augment current features.   These two sets are initialized to be empty sets in the begining of the FSS (see line \ref{FSS:initialization}).   Features are selected at each iteration which can approximately maximally reduce losses between current forecasts and actual sales over training set $\mathcal{D}^T$.   As such, parameters for current attributes csA and interactions csI are estimated first at line \ref{FSS:currentParametersEstimation} by the ABGD method.   Fitted sales $\hat{d}^{\textrm{cf}}_{is}$ for training observation $(i, s)$ is computed based on the estimated parameters  $\Theta^{\textrm{cf}}$ for current features at line \ref{FSS:computeFitted}.  Next depending on input search direction sd,  a subset set of either attributes or interactions which  are not in the current feature set are selected such that losses between forecasts $\hat{d}^{\textrm{cf}}_{is}$ and actual sales $d_{is}$  over all training observations $(i, s) \in T$ is reduced as much as possible.    Note that there are two components in the output of Algorithm \ref{alg:FS}: $\Delta A$ and $\Delta I$.  And at most one of them is non-empty.  
\begin{algorithm}[htbp]
	\begin{small}
		\caption{Forward Subset Selection} \label{alg:FS}
		\begin{algorithmic}[1]
			\item[//Inputs:]
			\item[//search direction: $\textrm{sd} \in \{1, -1\}$:]
			\item[//   $1$: forward attribute subset selection; $-1$: forward interaction subset selection]
			%\item[//current attriute subset, $\textr$] description
			\item[//csA, current attribute subset; csI, current interaction subset]
			\item[//loss function, $l(\cdot)$] 
			\item[//Outputs:]
			\item[//selected subsets of attributes and interactions for adding: $\{\Delta A, \Delta I\}$]
			\Procedure{FSS}{$\textrm{sd}$, $\textrm{csA}$, $\textrm{csI}$, $l(\cdot)$} 
			\State initialization, $\Delta A \gets \emptyset$,  $\Delta I \gets \emptyset$ \label{FSS:initialization}
			\State estimate parameters of current features, $\Theta^{\textrm{cf}} \gets \textrm{ABGD} (\mathcal{D}^{T}, \textrm{csA}, \textrm{csI}, l(\cdot))$\label{FSS:currentParametersEstimation} % \sigma, \eta, \{\lambda_{\theta} = 0 | \theta \in \Theta \})$ 
			\State for each training observation $(i, s) \in T$, compute its fitted sales with current features, $\hat{d}^{\textrm{cf}}_{is} = f(\mathbf{x}^{is}; \Theta^{\textrm{cf}})$\label{FSS:computeFitted}		
			\If{sd is positive one of forward attribute subset selection}
			\State for each out-of-current attribute $c \in A \backslash  \textrm{csA}$, compute its minimal loss $J^{*}_{\textrm{FAS},c}$ following equation (\ref{eq:FLS:minimalObjective})
			\State sort $\{J^{*}_{\textrm{FAS},c}| c \in A \backslash \textrm{csA}\}$ in ascending order, $J^{*}_{\textrm{FAS},c_{[1]}} \leq J^{*}_{\textrm{FAS}, c_{[2]}} \leq \cdots \leq J^{*}_{\textrm{FAS}, c_{[|A \backslash \textrm{csA}|]}}$\label{alg:fss:sort:levels:error} 
			\State  construct subset $\Delta A$ to be the top $\min\{b, |A \backslash \textrm{csA}|\}$ attributes in the above ranking list, $\Delta A \gets \{c_{[i]}| 1\leq i \leq \min\{b, |A \backslash  \textrm{csA} |\}\}$\label{alg:fss:select:levels}
			\Else
			\State for each out-of-current interaction $(c, c^{\prime}) \in I \backslash \textrm{csI}$, compute its minimal loss $J^{*}_{\textrm{FIS}, (c, c^{\prime})}$ following equation (\ref{eq:FIS:minimalObjective})
			\State sort $\{J^{*}_{\textrm{FIS}, (c, c^{\prime})} |(c, c^{\prime}) \in I \backslash \textrm{csI} \}$ in ascending order, $J^{*}_{\textrm{FIS}, (c, c^{\prime})_{[1]}} \leq J^{*}_{\textrm{FIS}, (c, c^{\prime})_{[2]}}  \leq \cdots \leq J^{*}_{\textrm{FIS}, (c, c^{\prime})_{[|I \backslash \textrm{csI} |]}}$\label{alg:fss:sort:interactions:error} 
			\State select the top $\min \{g, |I \backslash \textrm{csI}|\}$ interactions as $\Delta I$ from ranking list subject to two constraints:  (1) each two of them can not have common attributes and (2) every interaction can not have common attributes with existing interactions in csI \label{alg:fss:select:interactions}		 
			%\State select %\State select the top $\min \{g, |I \backslash \textrm{csI}|\}$ completely new interactions from the above ranking list, $\Delta I \gets \{(c, c^{\prime})_{[i]}| 1\leq i \leq \min\{g, |I \backslash  \textrm{csI} |\}, \}$
			%build subset $\Delta I$ to be the top $\min\{g, |I \backslash \textrm{csI}|\}$ interactions in the above ranking list, $\Delta I \gets \{(c, c^{\prime})_{[i]}| 1\leq i \leq \min\{g, |I \backslash  \textrm{csI} |\}\}$ \label{alg:fss:select:interactions}
			\EndIf
			\State \textbf{Return} $\{\Delta A, \Delta I\}$
			\EndProcedure
		\end{algorithmic}
	\end{small}
\end{algorithm}

%{\color{blue}
Next, we describe the details of how to select useful features. First, we discuss the case of input search direction sd being equal to positive one of forward attribute subset selection.  Considering an out-of-current attribute $c$ ($c \in A \backslash  \textrm{csA}$), adding it to the current attribute subset csA results  in new fitted values over training set $\mathcal{D}^T$. %}.  
Specifically,  for training observation $(i, s)$, we approximate the new fitted value to be 
$\hat{d}^{\textrm{cf}}_{is} \exp({\sum_{j \in L^{c}} \beta_{cj} x^{is}_{cj}})$ 
based on the assumption that  attributes are independent from each other. Adding attribute $c$ does not affect estimation of parameters for existing features.  Here $\{\beta_{cj} | j \in L^c\}$ are new parameters to be estimated. Adding attribute $c$ results in new losses as follows. 
\begin{equation} %\label{eq:FLS}
\mathcal{L}^{\textrm{NEW}} = 
\begin{cases}
\sum_{(i, s) \in T } [ \hat{d}^{\textrm{cf}}_{is} \exp({\sum_{j \in L^{c}} \beta_{cj} x^{is}_{cj}}) - d_{is}]^2 & \textrm{ if } l(\cdot) = l^{\textrm{ES}}(\cdot)\\ 
%\sum_{(i, s) \in T} (\hat{d}^{\textrm{cf}}_{is}  e^{\beta_j x^{is}_j} - d_{is})^2  & \textrm{ if } l(\cdot) = l^{\textrm{ES}}(\cdot)\\
\sum_{(i, s) \in T} [\frac{\hat{d}^{\textrm{cf}}_{is}}{d_{is}} \exp({\sum_{j \in L^{c}} \beta_{cj} x^{is}_{cj}}) - 1]^2  & \textrm{ if } l(\cdot) = l^{\textrm{PES}}(\cdot)
\end{cases}
\end{equation} 
Minimizing the above objective favors attributes with large number of levels, which produces good predictions over training set but is likely to overfit the data.  Therefore, we take the number of levels into account and define new objective $J_{\textrm{FAS}, c}$ as follows.  %($j \in L^c$) as follows. 
\begin{equation} \label{eq:FLS}
J_{\textrm{FAS},c} = 
\begin{cases}
\sum_{(i, s) \in T } [\hat{d}^{\textrm{cf}}_{is}  \exp({\sum_{j \in L^{c}} \beta_{cj} x^{is}_{cj}}) - d_{is}]^2  + \lambda_{\textrm{A,ES}} |L^c| & \textrm{ if } l(\cdot) = l^{\textrm{ES}}(\cdot)\\ 
%\sum_{(i, s) \in T} (\hat{d}^{\textrm{cf}}_{is}  e^{\beta_j x^{is}_j} - d_{is})^2&\textrm{ if } l(\cdot) = l^{\textrm{ES}}(\cdot)\\
\sum_{(i, s) \in T} [ \frac{\hat{d}^{\textrm{cf}}_{is}}{d_{is}} \exp({\sum_{j \in L^{c}} \beta_{cj} x^{is}_{cj}}) - 1]^2 + \lambda_{\textrm{A,PES}} |L^c| & \textrm{ if } l(\cdot) = l^{\textrm{PES}}(\cdot)
\end{cases}
\end{equation}  
Here $|L^c|$ represents the number of levels of attribute $c$; $\lambda_{\textrm{A,ES}}$ and $\lambda_{\textrm{A,PES}}$ are hyper-parameters and control the trade-off between the attribute's complexity and its fitting  over the training set.  Increasing their values leads to a price to pay for selecting attributes with a large number of levels.   Minimizing $J_{\textrm{FAS},c}$ is easily solved by setting the first derivate with respect to $e^{\beta_{cj}}$ to zero. Below is the minimal loss, $J^{*}_{\textrm{FAS},c}$, by attribute $c$ for the two loss functions. 

\begin{equation} \label{eq:FLS:minimalObjective}
	\begin{split}
		J^{*}_{\textrm{FAS}, c} & = \operatorname{Min}_{\{\beta_{cj} \in \mathbb{R}, j \in L^c\}} J_{\textrm{FAS},c} (\beta_{cj}, j \in L^c) \\
		 & =  
		 \begin{cases}
		 	\sum_{j \in L^c} \sum_{(i, s) \in T_{cj}} (\hat{d}^{\textrm{cf}}_{is}  w^{\textrm{ES}}_{cj} - d_{is})^2 + \lambda_{\textrm{A,ES}} |L^{c}|   & \textrm{ if } l(\cdot) = l^{\textrm{ES}}(\cdot) \\
		 	\sum_{j \in L^c} \sum_{(i, s) \in T_{cj}} (r_{is} w^{\textrm{PES}}_{cj} - 1)^2 + \lambda_{\textrm{A,PES}} |L^c|  & \textrm{ if } l(\cdot) = l^{\textrm{PES}}(\cdot)
		 \end{cases}
	\end{split}
\end{equation}
Here  $T_{cj} = \{(i, s)|x^{is}_{cj} = 1, (i, s) \in T \}$,  $w^{\textrm{ES}}_{cj} = \frac{\sum_{(i, s) \in T_{cj}} d_{is} \hat{d}^{\textrm{cf}}_{is} }{\sum_{(i, s) \in T_{cj}} (\hat{d}^{\textrm{cf}}_{is})^2} $,  $r_{is} = \frac{\hat{d}^{\textrm{cf}}_{is}}{d_{is}}$ and $w^{\textrm{PES}}_{cj} = \frac{\sum_{(i, s) \in T_{cj}} r_{is}}{\sum_{(i, s) \in T_{cj}} r^2_{is}}$.  Note that, for training observation $(i, s)\in T$ and level $j$ of attribute $c$, either $x^{is}_{cj} = 1$ or $x^{is}_{cj} = 0$ only must be true.  For level $j$ ($j \in L^c$),  set $T_{cj}$ contains all training observations associated with level $j$ of attribute $c$.   
We apply a greedy strategy and select the top $b$ attributes associated with the smallest $b$ minimal losses as result of $\Delta A$. Note that $b$ is a hyper-parameter and is denoted as the \textit{forward attribute selection depth}. 
If there are less than $b$ attributes out of current set $\textrm{csA}$, all of them are put into result subset $\Delta A$. These operations are presented at lines \ref{alg:fss:sort:levels:error} and \ref{alg:fss:select:levels} in Algorithm \ref{alg:FS}. 

Now we proceed to the case of input search direction sd being equal to negative one of forward interaction subset selection.  For an out-of-current interaction $(c, c^{\prime})$ ($(c, c^{\prime}) \in I \backslash \textrm{csI}$), it is possible that existing interactions in set $\textrm{csI}$  already contains  attribute $c$ or/and attribute $c^{\prime}$ already.   The parameters $\{\mu_{cj, p}\}$ (or/and $\{\mu_{c^{\prime}j^{\prime}, p}\}$) for attribute $c$ (or/and attribute $c^{\prime}$) associated with an interaction exists and have been estimated already.   Interactions with common attributes are highly correlated with each other.  Adding such interactions probably does not increase the predictive power % {\color{blue} 
\citep{cheng2014gradient}. For our case, we only take interactions into account if there are no common attributes.   We assume that these interactions are independent of each other. Adding new interaction $(c, c^{\prime})$ does not affect estimation of parameters for existing interactions.  We approximate the new fitted value to be 
$\hat{d}^{\textrm{cf}}_{is} \exp({\sum_{j \in L^c} \sum_{j^{\prime} \in L^{c^{\prime}}} x^{is}_{cj} x^{is}_{c^{\prime} j^{\prime}} \sum^f_{p=1} \mu_{cj,p} \mu_{c^{\prime} j^{\prime}, p}})$.  
In order to avoid over-fitting, we also take the interaction's complexity into account; which is the number of interaction levels. 
The objective FSS minimizes for searching interactions is denoted as $J_{\textrm{FIS}, (c, c^{\prime})}$  and defined as follows. 
\begin{equation} \label{eq:JFIS}
\begin{split}
& J_{\textrm{FIS}, (c, c^{\prime})}  = \\ %(\mathbf{\mu_{cj}}, \mathbf{\mu_{c^{\prime} j^{\prime}}})  = \\
& \begin{cases}
\sum_{(i, s) \in T} 
[\hat{d}^{\textrm{cf}}_{is} 
\exp({\sum_{j \in L^c} \sum_{j^{\prime} \in L^{c^{\prime}}} x^{is}_{cj} x^{is}_{c^{\prime} j^{\prime}} \sum^f_{p=1} \mu_{cj,p} \mu_{c^{\prime} j^{\prime}, p}})
 - d_{is}]^2 
+ \lambda_{\textrm{I,ES}} |L^c|\times |L^{c^{\prime}}| 
& \textrm{ if } l(\cdot) = l^{\textrm{ES}}(\cdot) \\
\sum_{(i, s) \in T} 
[\frac{\hat{d}^{\textrm{cf}}_{is}}{d_{is}}  \exp( {\sum_{j \in L^c} \sum_{j^{\prime} \in L^{c^{\prime}}} x^{is}_{cj} x^{is}_{c^{\prime} j^{\prime}} \sum^f_{p=1} \mu_{cj,p} \mu_{c^{\prime} j^{\prime}, p}} )
- 1]^2 + \lambda_{\textrm{I,PES}} |L^c|\times |L^{c^{\prime}}| 
& \textrm{ if } l(\cdot) = l^{\textrm{PES}}(\cdot)
\end{cases}
\end{split}
\end{equation} 
Here $\{\mu_{cj, p}|j \in L^c, p = 1, \ldots, f\}$ and $\{\mu_{c^{\prime} j^{\prime}, p} | j^{\prime} \in L^{c^{\prime}},  p = 1, \ldots, f\}$ are new parameters to be estimated.   Both $\lambda_{\textrm{I,ES}}$ and $ \lambda_{\textrm{I,PES}}$ are two hyper-parameters governing the complexity of interaction.     We observe that, by equations (\ref{eq:FLS}), (\ref{eq:JFIS}) and ignoring the penalty part for complexity,  minimizing  $J_{\textrm{FIS}, (c, c^{\prime})}$   is equivalent to minimizing  $J_{\textrm{FAS}, c \times c^{\prime}}$, %without considering the penalty part, 
where $c \times c^{\prime}$ is a synthetic attribute and defined as $x^{is}_{c \times c^{\prime}, j^{\prime \prime}} = x^{is}_{cj} \times x^{is}_{c^{\prime} j^{\prime}}$ and $j^{\prime \prime} = (j-1) \times |L^c| + j^{\prime}$.  %{\color{blue} 
The minimal value is denoted as $J^{*}_{\textrm{FIS}, (c, c^{\prime})}$ and provided as follows. %}. 
%defined as $x^{is}_{j \cap j^{\prime}} = x^{is}_{j} * x^{is}_{j^{\prime}}$ for training observation $(i, s) \in T$.  
%As such,  the minimal objective value of  $J_{\textrm{FIS}}(\mathbf{\mu}_j, \mathbf{\mu}_{j^{\prime}})$ can be derived in the following based on equation (\ref{eq:FLS:minimalObjective}). 
\begin{equation} \label{eq:FIS:minimalObjective}
\begin{split}
J^{*}_{\textrm{FIS}, (c, c^{\prime})} & = \operatorname{Min}_{\{\mathbf{\mu_{cj}}, \mathbf{\mu_{c^{\prime} j^{\prime}}} \in \mathbb{R}^f, j \in L^c, j^{\prime} \in L^{c^{\prime}} \}} J_{\textrm{FIS}, (c, c^{\prime})}(\mathbf{\mu_{cj}}, \mathbf{\mu_{c^{\prime} j^{\prime}}}) \\
&   = 
\begin{cases}
\sum_{j \in L^c} \sum_{j^{\prime} \in L^{c^{\prime}}} \sum_{(i,s) \in T_{cj, c^{\prime} j^{\prime}}} (\hat{d}^{\textrm{cf}}_{is} w^{\textrm{ES}}_{(c, c^{\prime})} - d_{is})^2 + \lambda_{\textrm{I,ES}} |L^{c}|\times |L^{c^{\prime}}| & \textrm{if } l(\cdot) = l^{\textrm{ES}}(\cdot) \\
\sum_{j \in L^c} \sum_{j^{\prime} \in L^{c^{\prime}}} \sum_{(i,s) \in T_{cj, c^{\prime} j^{\prime}}} (r_{is} w^{\textrm{PES}}_{(c, c^{\prime})} - 1)^2 + \lambda_{\textrm{I,PES}} |L^{c}|\times |L^{c^{\prime}}| & \textrm{if } l(\cdot) = l^{\textrm{PES}}(\cdot)
%\sum_{(i, s)\in　T^{j \cap j^{\prime}}} (\hat{d}^{\textrm{cf}}_{is} w^{\textrm{ES}}_{j \cap j^{\prime}} - d_{is})^2 + \sum_{(i, s) \in T \backslash T^{j \cap j^{\prime}}} (\hat{d}^{\textrm{cf}}_{is} - d_{is}) ^2   & \textrm{if } l(\cdot) = l^{\textrm{ES}}(\cdot) \\	%\sum_{(i, s) \in T^{j \cap j^{\prime}}} (r_{is} w^{\textrm{PES}}_{j \cap j^{\prime}} - 1)^2 + \sum_{(i, s) \in  T  \backslash  T^{j \cap j^{\prime}}}(r_{is} - 1)^2  & \textrm{if } l(\cdot) = l^{\textrm{PES}}(\cdot)
\end{cases}
\end{split}
\end{equation} 
Here  $T_{cj, c^{\prime} j^{\prime}} = \{(i, s) | x^{is}_{cj} = 1, x^{is}_{c^{\prime} j^{\prime}} = 1, (i, s) \in T \}$, $w^{\textrm{ES}}_{(c, c^{\prime})} = \frac{\sum_{(i, s) \in T_{cj, c^{\prime} j^{\prime}}} d_{is} \hat{d}^{\textrm{cf}}_{is} }{\sum_{(i, s) \in T_{cj, c^{\prime} j^{\prime}}} (\hat{d}^{\textrm{cf}}_{is})^2}$,  $r_{is} = \frac{\hat{d}^{\textrm{cf}}_{is}}{d_{is}}$ and $w^{\textrm{PES}}_{(c, c^{\prime})} = \frac{\sum_{(i, s) \in T_{cj, c^{\prime}j^{\prime}}} r_{is}}{\sum_{(i, s) \in T_{cj, c^{\prime}j^{\prime}}}r^2_{is}}$.  
%$T^{j \cap j^{\prime}} = \{(i, s)|x^{is}_j = 1,  x^{is}_{j^\prime} = 1, (i, s) \in T \}$,  $w^{\textrm{ES}}_{j \cap j^{\prime}} = \frac{\sum_{(i, s) \in T^{j \cap j^{\prime}}} d_{is} \hat{d}^{\textrm{cf}}_{is} }{\sum_{(i, s) \in T^{j \cap j^{\prime}}} (\hat{d}^{\textrm{cf}}_{is})^2}$,  $r_{is} = \frac{\hat{d}^{\textrm{cf}}_{is}}{d_{is}}$ and $w^{\textrm{PES}}_{j \cap j^{\prime}} = \frac{\sum_{(i, s) \in T^{j \cap j^{\prime}}} r_{is}}{\sum_{(i, s) \in T^{j \cap j^{\prime}}}r^2_{is}}$.  
Similarly to level subset selection, a greedy strategy is applied and the top $g$ interactions associated with the smallest $g$ minimal losses is built as result $\Delta I$.  And $g$ is a hyper-parameter and is denoted as the \textit{forward interaction selection depth}.  If there are less than $g$ interactions left, all of them are put into $\Delta I$ (see lines \ref{alg:fss:sort:interactions:error} and \ref{alg:fss:select:interactions}).

%However, we are not interested in estimating these new parameters.  Instead, %Instead, reduction on losses over training set $\mathcal{D}^{T}$ by this new interaction is our concentration. %has already formed %Given a new interaction $(i, j)$ ($(i, j) \in I \backslash \textrm{csI}$), %We assume that adding  %if the loss function is the PES loss, %If input search direction sd is positive one of forward level subset selection, %\subsubsection{Forward Level Subset Selection} \label{sec:fs:flss}  %The FLSS finds most useful levels  for augmenting current subset of levels $\textrm{csL}$. %\begin{algorithm} %\begin{small} %\caption{Forward Level Subset Selection} \label{alg:flss} %\begin{algorithmic}[1] %\Procedure{FLSS}{$\textrm{csL}$, $L$} %\State test %\State \textbf{Return} $\Delta L$ %\EndProcedure %\end{algorithmic} %\end{small} %\end{algorithm} %\subsubsection{Forward Interaction Subset Selection} \label{sec:fs:fiss}
\subsection{Overall Procedure} \label{sec:op}
In this section, we present the overall procedure (OP) which is outlined by Algorithm \ref{alg:OP}. It mainly consists of two stages: training and test stages.  

\begin{algorithm}[htbp]
	\begin{small}
		\caption{Overall Procedure} \label{alg:OP}
		\begin{algorithmic}[1]
			\item[//Inputs:]
			\item[//training set $\mathcal{D}^T$, test set $\mathcal{D}^E$, full attribute set $A$, full interaction set $I$]
			\item[//loss function $l(\cdot)$ of either the ES or the PES] 
			\item[//Outputs:]
			\item[//forecasts for training set $\{\hat{d}_{is} | (i, s) \in E\}$]
			\item[//MAPE, MAE and OR at both SKU-chain and SKU-store aggregate levels]  
			\Procedure{OP}{$\mathcal{D}^{T}$, $\mathcal{D}^{E}$, $I$, $l(\cdot)$} 
			\item[$~~$//Training stage:]
			\State select attribute subset  $\textrm{sA}^{*}$  and interaction subset $\textrm{sI}^{*}$ by GFSFS,  ($\textrm{sA}^{*}$,    $\textrm{sI}^{*}$)$\gets$GFSFS($\mathcal{D}^{T}$, $l(\cdot)$, $L$, $I$)\label{op:fs} 
			\State estimate parameters by ABGD,  $\Theta^{*}\gets $ABGD($\mathcal{D}^T$, $\textrm{sA}^{*}$, $\textrm{sI}^{*}$, $l(\cdot)$) \label{op:estimation}
			\item[]
			\item[$~~$//Test stage:]
			\State for each observation $(i, s) \in E$ in test set,  compute forecast $\hat{d}_{is} = f(\mathbf{x}^{is}; \Theta^{*})$  following equation (\ref{model:EFM})
			%\State calculate $\textrm{MAPE}^E_{\textrm{SKU-store}}$, $\textrm{MAPE}^E_{\textrm{SKU-chain}}$, $\textrm{MAE}^E_{\textrm{SKU-store}}$, $\textrm{MAE}^E_{\textrm{SKU-chain}}$, $\textrm{OR}^{E}_{\textrm{SKU-store}}$, $\textrm{OR}^{E}_{\textrm{SKU-chain}}$ following equations from (\ref{eq:mape:store}) to  (\ref{eq:or:chain}), respectively
			%\State \textbf{Return} $\{\{\hat{d}_{is} | (i, s) \in E\}$; MAPE, MAE and OR at both SKU-chain and SKU-store aggregate levels
			\State calculate $\textrm{MAPE}^E_{\textrm{SKU-store}}$, $\textrm{MAPE}^E_{\textrm{SKU-chain}}$, $\textrm{MAE}^E_{\textrm{SKU-store}}$, $\textrm{MAE}^E_{\textrm{SKU-chain}}$ following equations from (\ref{eq:mape:store}) to  (\ref{eq:mae:chain}), respectively
			\State \textbf{Return} $\{\{\hat{d}_{is} | (i, s) \in E\}$; MAPE and MAE for both SKU-store and SKU-chain sales forecasting
			%MAPE and MAE at both SKU-chain and SKU-store aggregate levels
			$\}$
			\EndProcedure
		\end{algorithmic}
	\end{small}
\end{algorithm}
%--------------------Up to Here----
In the training stage, features are selected first in the GFSFS algorithm.    The optimal attribute subset $\textrm{sA}^{*}$ and interaction subset $\textrm{sI}^{*}$ are produced (line \ref{op:fs} of Algorithm \ref{alg:OP}). Parameters for the selected attributes and interactions are then estimated by ABGD algorithm (line \ref{op:estimation} of Algorithm \ref{alg:OP}).  %Parameters for selected attributes and interactions %associated with these features %are estimated by ABGD algorithm.   
Note that in the process of feature selection, parameters are needed to be estimated and ABGD is called in CV and FSS. %{\color{blue} 
There exist differences between these two estimations in terms of the value of regularization hyper-parameter $\lambda_{\theta}$. %}.  In 
For feature selection process (line \ref{op:fs}), all regularization hyper-parameters $\{\lambda_{\theta}\}$  in the ABGD are set to zero %because, at this step, using ABGD 
to estimate parameters %is 
to ascertain %get 
the importance ranking for attributes and interactions.  At the same time, we have also considered  %and we have considered already 
the over-fitting problem in feature importance measures in equations (\ref{eq:FLS}) and (\ref{eq:JFIS}).   However, after features are selected, we re-estimate parameters with the whole training set and they will be used for produce forecasts for the test set.  At this step, we should take the regularization into account to avoid over-fitting. %{\color{blue}  %Settings of regularization hyper-parameters $\{\lambda_{\theta}\}$ are discussed in Section \ref{sec:com:settingsOfHyperParam}. %}.
%Note that, %After features are selected, they should be re-estimated.  In the CV, only a part of training data is used for estimating them. %and regularization is ignored. %The purpose of those estimation  is to get important information for finding useful and relevant attributes and interactions.   %After they are selected,  all training data is used for re-estimating parameters for selected features.  And finally these estimated parameters are used to produce forecasts for the hold-out test set $\mathcal{D}^E$. %And then these estimated parameters are used to produce forecasts for the hold-out test set $\mathcal{D}^E$ in the end. %And they are used to produce forecasts for the hold-out test set $\mathcal{D}^E$.

In test stage, %{\color{blue} 
forecasts for test set $\{ \hat{d}_{is} | (i, s) \in E \}$ are calculated based on the estimated parameters for the EFM model at the training stage. %}.  
Then we evaluate the performance and compute two indicators: mean absolute percentage error (MAPE) and mean absolute error (MAE). 
%and overestimation ratio (OR).  %Note that 
Forecast $\hat{d}_{is}$ is for SKU $i$ at store $s$.  
It is at the aggregation level of SKU-store. As discussed in Section \ref{intro}, forecasts are aimed at facilitating the buying quantity decisions which are usually an aggregate decision for all stores instead of individual stores. Therefore, the SKU-store sales forecasts are consolidated to forecasts for all stores (SKU-chain forecast) and then used as the buying quantity for each SKU.
% {\color{blue} It  is  at  the aggregation level of SKU-store. As discussed in Section \ref{intro}, forecasts are aimed at facilitating initial buy quantity decisions  which  are usually  an  aggregate  decision  for all stores  instead  of for each  individual  one. Therefore, SKU-store sales forecasts are consolidated to forecasts for the SKU over the chain (all stores) (SKU-chain forecast) and then used as initial buy quantity for each SKU}. 
% {\color{blue} It is the SKU-store sales quantity.  We note that forecasts are aimed at facilitating initial buy quantity decision which is usually an aggregate decision for all stores instead of for each individual one.  %assisting initial buy %Note that,  %Note that, forecasts in current study are  %Forecasts  $\{\hat{d}_{is} | (i, s) \in E \}$ are aimed at using as initial buy quantities,  
%Current forecasts $\{\hat{d}_{is} | (i, s) \in E \}$ are used as initial buy quantities.  The initial buy quantity  %which is usually for a SKU over  all stores (the whole chain) and noted as SKU-chain forecast.   %As such, forecasts at the aggregation level of SKU-store are consolidated to the aggregation level of SKU-chain and then used for making initial order.  %Therefore,  SKU-store sales forecasts %at the aggregation level of SKU-store  %are consolidated to %the aggregation level of  %SKU-chain forecast and then used as initial buy quantity for each SKU. %As such, 
As such, we calculate both MAPE and MAE %and %OR 
for both SKU-store and SKU-chain sales forecasting. %at the aggregation level of SKU-chain as well.  
Given SKU-store forecasts $\{\hat{d}_{is} | (i, s) \in E \}$ and actual sales $\{d_{is} | (i, s) \in E \}$,  formulations of the MAPE and MAE %and OR 
for %at aggregation levels of 
both SKU-store and SKU-chain forecasting are given as follows.  We denote all SKUs in test set $E$ as %$E^{\textrm{SKU}}$, 
$E^{\textrm{SKU}} = \{ i | \exists s \in M \textrm{ such that } (i, s) \in E \}$.
\begin{itemize}
	\item MAPE
	\begin{equation} \label{eq:mape:store}
	\textrm{MAPE}^{E}_{\textrm{SKU-store}} = \frac{1}{|E|} \sum_{(i, s) \in E}  |\frac{\hat{d}_{is} - d_{is}}{d_{is}}|
	\end{equation}
	\begin{equation}
	\textrm{MAPE}^{E}_{\textrm{SKU-chain}} = \frac{1}{|E^{\textrm{SKU}}|} \sum_{i \in E^{\textrm{SKU}}} |\frac{\sum_{s\in M} \hat{d}_{is} - \sum_{s \in M} d_{is}}{\sum_{s \in M} d_{is}}|
	\end{equation}
	\item MAE
	\begin{equation}
	\textrm{MAE}^{E}_{\textrm{SKU-store}} = \frac{1}{|E|} \sum_{(i, s) \in E} |\hat{d}_{is} - d_{is}| 
	\end{equation}
	\begin{equation}\label{eq:mae:chain}
	\textrm{MAE}^{E}_{\textrm{SKU-chain}} = \frac{1}{|E^{\textrm{SKU}}|} \sum_{i \in E^{\textrm{SKU}}} |\sum_{s\in M} \hat{d}_{is} - \sum_{s \in M} d_{is}|
	\end{equation}
	%	\item OR
	%		\begin{equation}
	%			\textrm{OR}^{E}_{\textrm{SKU-store}} = \frac{1}{|E|} \sum_{(i, s) \in E} I_{\hat{d}_{is} \geq d_{is}}
	%		\end{equation}
	%		\begin{equation} \label{eq:or:chain}
	%			\textrm{OR}^E_{\textrm{SKU-chain}} = \frac{1}{E^{\textrm{SKU}}} \sum_{i \in E^{\textrm{SKU}}} I_{(\sum_{s\in M, d_{is} > 0} \hat{d}_{is}) \geq (\sum_{s \in M, d_{is} > 0} d_{is})}  
	%		\end{equation}
	%	Here $I_{\hat{d}_{is} \geq d_{is}}$ is the indicator function and takes value of 1 if $\hat{d}_{is} \geq d_{is}$; 0 otherwise. 
\end{itemize}

\section{Computational Studies}  \label{sec:cor} 
In this section,  we present the computational studies including not only  the sales forecasting dataset but also two public datasets.  Several important variants of the EFM model are investigated.  Comparisons and insights are provided.    Note that our proposed algorithms are implemented in Java.   Hardware configurations for these studies are given as follows.   Computations for hyper-parameter searches are performed on Supermicro Linux server with 4-core Intel Xeon E5-2623V3 (10MB L3, 3.0GHz), 256 RAM and Ubuntu 16.04 operating system.    The remaining implementations are conducted on a  Dell personal computer with an Intel Core i7-6700 CPU, 3.40 GHz, 32 GB RAM and 64-bit Windows 10 operating system.  We investigate different variants and extensions of the proposed EFM model for comparion  analysis.  For ease of presentation, we denote the method as follows.   
\begin{itemize}
\item \textbf{EFM-PES}: the proposed  EFM model with PES loss function and the PES  based feature selection.  Hyperparameters were got via the grid search.   For this method,   we denote the selected attribute set as $\textrm{sA}^*_{\textrm{PES}}$  and the  selected attribute interaction set as $\textrm{sI}^*_{\textrm{PES}}$.   	

\item \textbf{EFM-ES}:  the proposed method with ES loss function and the ES based feature selection.   Hyper-parameters are set by the grid search.   Similarly,  we denote the selected attribute set  as $\textrm{sA}^*_{\textrm{ES}}$ and the selected attribute interaction set as $\textrm{sI}^*_{\textrm{ES}}$.  		

\item \textbf{logFM-PES}:   this method first performs a  log  transform of training data $\mathcal{D}^T = \{(\mathbf{x}^{is},d_{is}) | (i, s) \in T \}$  into  $\mathcal{D}^T_{\textrm{LOG}} = \{(\mathbf{x}^{is}, \log(d_{is})) | (i, s) \in T \}$.  	Then $\mathcal{D}^T_{\textrm{LOG}} $ is used to train the original FM model with attribute set $\textrm{sA}^*_{\textrm{PES}}$,  attribute interaction set  $\textrm{sI}^*_{\textrm{PES}}$ and PES loss by the ALS method \cite{rendle2011fast}.   The fitted model $f^{\textrm{logFM}}(\cdot, \Theta)$ is used to forecast for the test set  $\mathcal{D}^E = \{(\mathbf{x}^{is}, d_{is}) | (i, s) \in E \}$,   and the final prediction is computed as $\hat{d}^{\textrm{logFM}}_{is} = \exp(f^{\textrm{logFM}}(\mathbf{x}^{is}, \Theta))$ for  any  $(i,s) \in E$ which is used to compute performance indicators of MAPE and MAE.   

\item \textbf{logFM-ES}:  this method is similar to the above logFM-PES with two differences: (1) the loss function is ES and (2) attributes and attribute interactions are $\textrm{sA}^*_{\textrm{ES}}$  and $\textrm{sI}^*_{\textrm{ES}}$, respectively, by the EFM-ES method.  
\end{itemize}

\subsection{Retail  Sales Forecasting} \label{sec:sf}
\subsubsection{Database Introduction and Preprocessing}  \label{sec:dataIntro}
The database is provided by our industry partner, the retailer of ladies footwear in Singapore.  The raw data mainly consists of two parts: (1) attributes associated with each SKU and (2) sales transactions from the point-of-sales (POS) system. We first merge these two sources into one set (table).  Consequently, there are 45 explanatory attributes (variables) and 1 response variable.  The raw data is preprocessed  and they include removal of  errors and outliers, filling missing values,  data anonymization without losing its nature,  consolidation as well as discretization.   The first three operations  follow  the standard data mining techniques while the consolidation refers to  aggregating fine-grained information  of each sales transaction over the  first eleven weeks after launch. The last preprocessing is data discretization.  In the proposed approach, we only take categorical attributes into account.   We employ an unsupervised data discretization of equal frequencies and domain knowledge to transform the numerical (continuous) attributes into categorical attributes.   The off-the-shelf function \textit{discretization} in \textit{infotheo} \citep{meyer1infotheo}  package in R is used here.  

\begin{figure}[htbp]
	\begin{center}
		\begin{tikzpicture}[scale = 0.65]
		\draw[green, line width = 12] (-10,0) -- (-1,0);
		\draw[thick, <-] (-10, 0.4) -- (-10, 2.5);
		\node[above] at (-10, 2.5) {1 January 2012};
		\draw[thick, <-] (-1, 0.4) -- (-1, 1); 	
		\node[above] at (-1.3, 1) {14 April 2013};
		\draw[yellow, line width = 12] (-1,0) -- (0.5, 0);
		\draw[thick, <-]  (0.5, 0.4) -- (0.5, 2.0);
		\node[above] at (0.5, 2.0) {30 June 2013};
		\draw[red, line width = 12] (0.5, 0) -- (1.3, 0);
		%\draw[thick, <-] (1.3, 0.4) -- (1.3, 1);
		%\node[above] at (2.1, 1) {31 July 2013}; 
		\draw[red, line width = 12] (1.3, 0) -- (4.5, 0);
		\draw[thick, <-] (4.5, 0.4) -- (4.5, 2.0);
		\node[above] at (4.5, 2.0) {31 December 2013};
		\draw[green, line width = 12] (4.5, 0) -- (7.5, 0);	
		\draw[thick, <-] (7.5, 0.4) -- (7.5, 1);
		\node[above] at (7.5, 1) {3 May 2014};
		\draw[yellow, line width = 12] (7.5, 0) -- (9, 0);
		\draw[thick, <-] (9, 0.4) -- (9, 2.5);
		\node[above] at (9, 2.5) {20 July 2014};	
		
		\draw[thick] (-10, -0.4) -- (-10, -2.5);
		\draw[thick] (-1, -0.4) -- (-1,-1.5);
		\draw[thick] (0.5, -0.4) -- (0.5, -2.5);
		%\draw[thick] (1.3, -0.4) -- (1.3, -2.5);
		\draw[thick] (4.5, -0.4) -- (4.5, -2.5);
		\draw[thick] (7.5, -0.4) -- (7.5,-1.5);
		\draw[thick] (9, -0.4) -- (9, -2.5);	
		
		\node at (-4.75, -2) {training};
		\draw[thick,  <-] (-10, -2) -- (-6, -2);
		\draw[thick, ->] (-3.5, -2) -- (0.5, -2);
		
		\node at (6.75, -2) {test};
		\draw[thick,  <-] (4.5, -2) -- (5.75, -2);
		\draw[thick, ->] (7.75, -2) -- (9, -2);
		
		\node at (2.5, -2) {lead time};
		\node at (2.5, -2.5) {(six months)};
		\draw[thick,  <-] (0.5, -2) -- (1.3, -2);
		\draw[thick, ->] (3.7, -2) -- (4.5, -2);
		
		\node at (-5.5, -1) {launch dates};
		\draw[thick, <-] (-10, -1) -- (-7, -1); 
		\draw[thick, ->] (-4, -1) -- (-1, -1);
		
		\node[below] at (-0.85, -3){product life};
		\node[below] at (-0.85, -3.5) {(eleven weeks)};
		\draw[thick, <->] (-1, -1) -- (0.5, -1);
		\draw[thick, ->] (-0.25, -3) -- (-0.25, -1);
		
		\node at (6, -1) {launch dates};
		\node[below] at (8.25, -3){product life};
		\node[below] at (8.25, -3.5) {(eleven weeks)};
		\draw[thick, <->] (7.5, -1) -- (9, -1);
		\draw[thick, ->] (8.25, -3) -- (8.25, -1);

		\end{tikzpicture} 
	\end{center}
	\caption{Partition of data into training and test sets} \label{Fig:trainingvstest}
\end{figure}
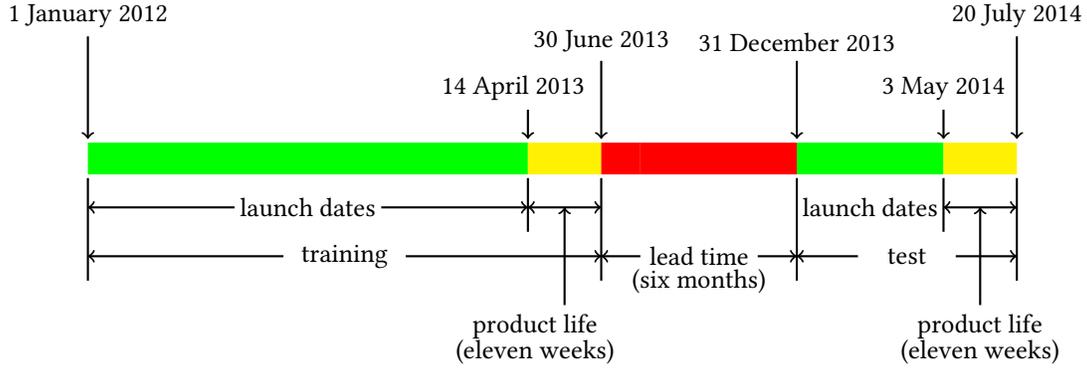

Our database covers the sales occurring in retailer's stores in Singapore for the time period from 1 January 2012 to 20 July 2014.  We divide the data into training and test sets. Observations associated with launch time from 1 January  2012 to 14 April 2013 are constructed as training set $\mathcal{D}^T$. And those with launch time from 31 December 2013 to 3 May 2014 are composed as test set $\mathcal{D}^E$.  The rationale is illustrated by Figure \ref{Fig:trainingvstest}. We take both lead time of six month and product life time of 11 weeks into account. We apply the proposed solution on three classes  with the masked names of 69-y6, p-1 and x-9w, respectively.  Table \ref{tab:descriptiveStatistics} lists sizes of training and test for the three classes.

\begin{table}[htbp]
	%\begin{small}
		\caption{Sizes of training and test for 3 classes}  \label{tab:descriptiveStatistics}
		\begin{tabular}{ccc}    
				\toprule
				Class (masked name) &	Training $\mathcal{D}^T$ & Test  $\mathcal{D}^E$ \\ %& Levels $L$  & Interactions $I$\\
				\hline
				69-y6 & 4,984 & 1,179 \\  %& 258 & 12,953 \\
				p-1   & 6,222 & 1,824 \\  % & 301 & 16,519 \\
				x-9w  & 5,526 & 1,481 \\  %& 267 & 14,180 \\
				\hline
		\end{tabular}
	%\end{small}
\end{table}

\subsubsection{Methods and Settings}   Besides EFM-PES, EFM-ES, logFM-PES and logFM-ES, we also applied the following  methods for  current sales forecasting application.  

\begin{itemize}
	\item \textbf{SP-EFM-PES}:  this method takes the sales quantity of similar SKUs (products) sold at the same store  into account with an extended EFM model  which is formulated as  follows.  	
		 \begin{equation} \label{model:GFM-e}
		 \begin{split}
		 \hat{d}^{\textrm{E}}_{is} = & f^{\textrm{E}}(\mathbf{x}^{is}; \Theta) =    \exp(\beta_0  + \sum_{c \in A} \sum_{j \in L^c} \beta_{cj} x^{is}_{cj}     +   \sum_{b \in B}  \beta_b z^{is}_b   + \sum_{(c,  c^{\prime}) \in I} \sum_{j \in L^c} \sum_{j^{\prime} \in  L^{c^{\prime}}}  x^{is}_{cj} x^{is}_{c^{\prime} j^{\prime}} <\mathbf{\mu_{cj}}, ~\mathbf{\mu_{c^{\prime}  j^{\prime}}}>  \\
		 &  + \sum_{c  \in A}  \sum_{j \in L^c}  \sum_{b  \in B}  x^{is}_{cj} z^{is}_b <\gamma_{cj}, \gamma_b>   + 
		 \sum_{b  \in B}   \sum_{b^{\prime} < b,  b^{\prime}  \in B}   z^{is}_b  z^{is}_{b^{\prime}}  <\zeta_{b}, \zeta_{b^{\prime}}>)
		 \end{split}
		 \end{equation} 
	Here $<\gamma_{cj}, \gamma_b> = \sum^r_{p=1} \gamma_{cj, p}  \gamma_{b,p}$ models the intersection between continuous variable $z^{is}_b$ and categorical variable $x^{is}_{cj}$;  $<\zeta_{b}, \zeta_{b^{\prime}}> = \sum^{t}_{p=1 } \zeta_{b,p}  \zeta_{b^{\prime}, p}$ considers the interactions between two continuous variables $z^{is}_b$ and $z^{is}_{b^{\prime}}$;  $B$ is the index set of continuous variables.   We first define the similarity between two SKUs before presenting the detail of $B$  and $z^{is}_b$.  	For SKU $i$ and SKU $i^{\prime}$  at store $s$,   the  similarity  between them is   
	\begin{equation}
	\textrm{Similarity}^s_{i, i^{\prime}} = \frac{\sum_{c \in \{ \textrm{sA}^{*}_{\textrm{PES}} \cup A^{\textrm{sI}^{*}_{\textrm{PES}}} \} } \sum_{j \in L^c}  \mathbb{1} (x^{is}_{cj} == x^{i^{\prime} s}_{cj})  }
	{|\textrm{sA}^{*}_{\textrm{PES}} \cup A^{\textrm{sI}^{*}_{\textrm{PES}}} |}. 
	\end{equation}	
	Where  $A^{\textrm{sI}^{*}_{\textrm{PES}}}$ is the attribute set associated with intersection $\textrm{sI}^{*}_{\textrm{PES}}$ which is $A^{\textrm{sI}^{*}_{\textrm{PES}}}  = \{ c  | \exists c^{\prime} \in A \textrm{ such that } (c, c^{\prime}) \in  \textrm{sI}^{*}_{\textrm{PES}} \textrm{ or } (c^{\prime}, c) \in \textrm{sI}^{*}_{\textrm{PES}} \}$,   and  $\mathbb{1}(x) = 1$ if $x$ is True; 0 otherwise.   
		%The continuous variables set $B$ includes 	%
	These continuous variables are  sales quantities of %the top-$K$ ($K = |B|$) 
	similar SKUs which have been sold for elven weeks at the  same store, specifically, for $b  \in B = \{1, \ldots, |B|\}$,   $z^{is}_b$ is the sales quantity of the top $b$ similar SKU in terms of the defined similarity to SKU $i$ at store $s$.   	
	%The continuous attribute $z^{is}_b$ is computed as $z^{is}_b = d_{bs} * S_{b, is}$,  and  $d_{bs}$ is the actual sales  of  SKU  $b$.  
	This extended model follows the linearity and the proposed learning method is  also applicable here.   The loss function is the PES.   The input attribute and attribute intersection  set for this model are  $\textrm{sA}^{*}_{\textrm{PES}}$  and $\textrm{sI}^{*}_{\textrm{PES}}$, respectively, by the EFM-PES method.  The set $B$   is set to be  top three similar sales quantities.  Interactions between continuous and categorical variables  are selected by  applying similar feature selection method with the same idea of the proposed  approach  in Section \ref{sec:fs} for categorical interactions.  We finally note that  this method is named as, SP-EFP-PES,   where ``SP'' %of whi
	is short for ``similar products''.    
	
	\item \textbf{SP-EFM-ES}:  this method is similar to the above SP-EFM-PES with two differences: (1) the loss function is ES and (2) the input attributes and attribute interactions are changed to  $\textrm{sA}^*_{\textrm{ES}}$  and $\textrm{sI}^*_{\textrm{ES}}$, respectively, by the EFM-ES method.   
	
	\item \textbf{Lasso}:   off-the-shelf model, glmnet(), from R package glmnet.  The input  features are  all  available attributes and attribute interactions, and the function's own feature selection mechanism is applied.  
	
	\item \textbf{Random Forest (RF)}:   off-the-shelf  model,   h2o.randForest(),  from R package h2o.   The model selects  features  automatically and  all available features are fed.   
	
	\item \textbf{Regression Tree (RT)}:   off-the-shelf model, rpart(), from R package rpart.  The features are selected by the function from all available features.  
	
	\item \textbf{Support vector regression (SVR)}:  off-the-shelf model,  svm(), from R package e1071.   All available features are fed into the function and it selects useful features first.     
\end{itemize}

\begin{table}[htbp]
		\centering
		%\begin{small}
		\caption{Sales forecasting  results  of MAPE and MAE for the three classes} \label{tab:forectR}
		\resizebox{1.0\columnwidth}{!}{
			\begin{tabular}{c ccc  ccc  ccc  ccc}
				\toprule	\midrule
				& \multicolumn{6}{c}{SKU-Store}                 & \multicolumn{6}{c}{SKU-Chain} \\	
				\cmidrule(lr{1em}){2-7} \cmidrule(rl{1em}){8-13}
				& \multicolumn{3}{c}{MAPE} & \multicolumn{3}{c}{MAE} & \multicolumn{3}{c}{MAPE} & \multicolumn{3}{c}{MAE} \\
				\cmidrule(lr{1em}){2-4} \cmidrule(rl{1em}){5-7} \cmidrule(rl{1em}){8-10} \cmidrule(rl{1em}){11-13}
				& \multicolumn{1}{c}{69-y6} & \multicolumn{1}{c}{p-1} & \multicolumn{1}{c}{x-9w} & \multicolumn{1}{c}{69-y6} & \multicolumn{1}{c}{p-1} & \multicolumn{1}{c}{x-9w} & \multicolumn{1}{c}{69-y6} & \multicolumn{1}{c}{p-1} & \multicolumn{1}{c}{x-9w} & \multicolumn{1}{c}{69-y6} & \multicolumn{1}{c}{p-1} & \multicolumn{1}{c}{x-9w} \\
				\midrule
				
				\textbf{EFM-PES (proposed)} & \textbf{4.55\%} & \textbf{5.55\%} & \textbf{5.42\%} & \textbf{1.59}  & 3.32  & 3.44  & \textbf{4.57\%} & \textbf{5.55\%} & \textbf{5.57\%} & 3.81  & 6.37  & 6.91 \\
				
				\textbf{EFM-ES (proposed)} & 5.15\% & 7.97\% & 8.14\% &  2.49 & \textbf{3.01} & \textbf{3.22} & 5.32\% & 8.23\% & 9.25\% & \textbf{3.61} & \textbf{5.41} & \textbf{5.25} \\
				
				logFM-PES & 6.90\% & 8.00\% & 10.00\% & 3.25  & 4.5   & 5.5   & 6.50\% & 8.52\% & 11.25\% & 4.25  & 8.5   & 7.5 \\
				logFM-ES & 7.20\% & 8.50\% & 12.00\% & 2.48  & 3.4   & 4.9   & 7.02\% & 8.96\% & 13.00\% & 3.91  & 7.2   & 6.5 \\
				SP-EFM-PES & 6.50\% & 8.36\% & 8.90\% & 3.62  & 5.5   & 5.83  & 8.20\% & 10.25\% & 8.80\% & 5.06  & 8.8   & 7.56 \\
				SP-EFM-E-ES & 7.23\% & 9.58\% & 7.90\% & 2.3   & 4.8   & 4.59  & 9.50\% & 11.50\% & 9.58\% & 4.05  & 7.8   & 6.59 \\
				Lasso & 85.99\% & 91.27\% & 90.24\% & 28.01 & 49.69 & 53.92 & 86.40\% & 91.77\% & 90.63\% & 67.95 & 98.2  & 111.52 \\
				RF  & 116.43\% & 30.18\% & 114.49\% & 31.6  & 14.26 & 47.82 & 109.67\% & 29.28\% & 111.01\% & 76.65 & 26.69 & 98.24 \\
				RT & 142.82\% & 72.04\% & 121.04\% & 39.03 & 31.54 & 51.05 & 135.05\% & 96.51\% & 117.29\% & 94.61 & 60.69 & 104.16 \\
				SVR   & 89.77\% & 47.75\% & 102.60\% & 24.56 & 20.35 & 43.97 & 96.40\% & 42.84\% & 96.24\% & 59.58 & 39.22 & 90.42 \\
				\midrule
				\bottomrule
			\end{tabular}%
		} %\end{small}
\end{table}%

\subsubsection{Results}  In this subsection, we present forecasting results for the three classes of ladies' footware: 69-y6, p-1 and x-9w.    %Table \ref{tab:forectR}  displays the results. 
Table \ref{tab:forectR} displays the results, from which  we can conclude that 
%,  we have several observations: 
(1) the EFM-PES and EFM-ES performs best for most cases in  terms  of MAPE and  MAE, respectively;  (2) the feature selection method proposed in current study is  effective because other feature selection methods (i.e., Lasso, RF,  RT,  SVR) result in unacceptable performance; (3) The logFM-PES and logFM-ES cannot perform better than EFM-PES and EFM-ES.    %This indicates that using log-transformation with the original FM is not that effective comparing with the proposed solution.   This is likely because  log-transformation has changed the variance of the response variable $d_{is}$ and results in an inefficient training/fitting process.     
We also note that our results  of EFM-PES and EFM-ES dominate results of retail literature.  The MAPEs for SKU-chain forecasting are around 5\% and MAEs are less than 7 units for either each individual class or overall of three classes.    It compares favorably with existing studies of sales/demand forecasting \citep{ferreira2015analytics, fisher2014demand, lee2003bayesian, fader1996modeling} among which the best reported  performance in terms of  MAPE is 16.2\% for a single new SKU by Fisher and Vaidyanathan (2014) \cite{fisher2014demand}.   For SKU-store forecasts, the overall performance is promising as well. %}.  

\subsection{Student  Performance and Burned Area of Fires Forecasting}
\subsubsection{Database Introduction and Preprocessing}   Here we further apply the proposed methods on two external  public datasets and investigate  the differences between EFM and logFM,  PES and ES loss functions.     These two public datasets are: 
\begin{itemize}
	\item \textbf{Secondary school student performance (SCTP)}  \cite{cortez2008using}:   this dataset is public available at UIC Machine Learning Repository\footnote{https://archive.ics.uci.edu/ml/datasets/student+performance} and Kaggle \footnote{https://www.kaggle.com/dipam7/student-grade-prediction},   and includes social,  demographic and school related  attributes and  Mathematics and Portuguese language grades of three periods: G1,  G2 and G3 (the  final grade).  There are 15 continuous attributes  and 17  categorical attributes; and data sizes are  395 and 649 for Mathematics (SCTP-M) and Portuguese (SCTP-P), respectively.   Here we use  this  dataset  to predict G3 the range of which is from 0  to 20.  In order  to compute MAPE, zeros of G3s are revised to 0.1  in data preprocessing step.    
	
	\item \textbf{Forest fires (FF)} \cite{cortez2007data}: this dataset  is also public available at UIC Machine Learning Repository\footnote{https://archive.ics.uci.edu/ml/datasets/Forest+Fires} and Kaggle \footnote{https://www.kaggle.com/elikplim/forest-fires-data-set}.    It is on forest fires including meteorological information and can be used to predict the burned area.  There are 12 explanatory attributes including 2 categorical and 10 numeric.   And there are 517 instances.  In order to compute MAPE,  zero values of  response variable,  \textit{burned area}, are preprocessed as 0.1.   
	%There are  %\item \textbf{House prices (HP)}:  
\end{itemize}

\begin{table}[htbp]
	%\begin{small}
	\caption{Hyper-parameters  for FM based methods for SCTP and FF  Dataset} \label{tab:hyperParam}
	\resizebox{0.8\columnwidth}{!}{
		\begin{tabular}{clccc}    
			\toprule
			\multirow{2}{*}{Loss function} & \multirow{2}{*}{Hyper-parameter} & \multicolumn{3}{c}{Dataset}\\
			\cline{3-5}
			& & SCTP-P & SCTP-M & FF \\
			\hline
			\multirow{10}{*}{PES} & learning rate $\eta$ & $ 4.95\times 10^{-6}$ &  $ 3.5\times 10^{-6}$ & $1.95\times 10^{-6}$  \\
			& maxInteractions & 4000 &  4000 & 15000 \\
			& regularization for all attribute selection $\lambda_{\textrm{A,PES}}$ & 0.005 & $1.0 \times 10^{-3}$ & $5.0 \times 10^{-4}$\\
			& regularization for all interaction selection $\lambda_{\textrm{I,PES}}$ & 0.10 & $1.0 \times 10^{-3}$ & $5.0 \times 10^{-4}$\\
			& regularization for all attributes $\lambda_v$ & $1\times 10^{-3}$ &  0.1 & 0.1 \\
			& regularization for all interactions $\lambda_w$  & 10 &  0 & 0 \\
			&  standard  deviation $\sigma$ (initialization) &  0.1 & 0.1 & 0.1 \\
			& attribute selection depth $b$ (categorical and numeric)  &  3 & 3 & 2 \\
			& interaction selection depth $g$ (categorical and numeric)   & 2 & 2 & 1 \\
			& factorization dimensionalities $f, r, t$ (categorical and numeric)  &   2 & 2  & 2 \\
			\hline
			\multirow{10}{*}{ES} & learning rate $\eta$ & $4.80\times 10^{-10}$ &  $3.15\times 10^{-10}$ & $2.05 \times 10^{-10}$  \\
			& maxInteractions & 5000 &  10000 & 17000 \\
			& regularization for all attribute selection $\lambda_{\textrm{A,ES}}$ & 1000 & 100 & 100 \\
			& regularization for all interaction selection $\lambda_{\textrm{I,ES}}$ & 1000 & 100 & 500 \\
			& regularization for all attributes $\lambda_v$ & 100 &  0  & 0 \\
			& regularization for all interactions $\lambda_w$  & 0  & 0  & 0   \\
			&  standard  deviation $\sigma$  (initialization) &  0.1 & 0.1 & 0.1 \\
			& attribute selection depth $b$ (categorical and numeric)  &  3 & 3 & 2 \\
			&  interaction selection depth $g$ (categorical and numeric)  & 2 & 2 & 1 \\
			& factorization dimensionalities $f, r, t$ (categorical and numeric)  &   2 & 2  & 2 \\
			\bottomrule
		\end{tabular}
	}
	%\end{small}
\end{table}

\subsubsection{Methods and  Settings}   We apply EFM-PES, EFM-ES, logFM-PES, logFM-ES,  SVR and RF.   Here EFM-PES and EFM-ES are the extended EFM model (\ref{model:GFM-e}) with numeric variables.  The evaluation metrics used are MAPE and MAE.   In order to measure the predictive power of these methods,  five-fold cross-validation is applied here;   each dataset is randomly divided into five parts of  equal  size; and then one subset is used as test and the remaining is used as training; five runs are performed.  Finally, average  MAPE  and average MAE of test  set over the five runs are computed as the results.   Now we present settings of hyper-parameters and features.  The settings of SVR  and RF follow the best results reported by  Cortez et al. \cite{cortez2008using} and Cortez et al.  \cite{cortez2007data} for SCTP  and FF datasets, respectively.   For the remaining FM based methods (i.e., EFM-PES, EFM-ES, logFM-PES and logFM-ES),   features are selected by an extended approach from Section \ref{sec:fs} with the capability of selecting continuous  variables; the setting between exponential formulations (i.e., EFM-PES and EFM-ES) and log-transformations (i.e., logFM-PES and logFM-ES) are similar to that of sales forecasting in Section \ref{sec:sf} .   Hyper-parameters are given in  Table \ref{tab:hyperParam} and most of them (i.e., learning rate, maxInteractions, regularization parameters) are got by grid search.% and the rest is set by rule of thumb .     %In the end we note  that all implementations are conducted on a  Dell personal computer with an Intel Core i7-6700 CPU, 3.40 GHz, 32 GB RAM and 64-bit Windows 10 operating system. 

\subsubsection{Results} Table \ref{tab:forectRE} displays the results and 
%From Table \ref{tab:forectRE}, 
we have several observations:  (1) the performance of FM based methods (i.e., EFM-PES, EFM-ES,  logFM-PES and logFM-ES) are better than that of SVR and RF; (2) the EFM-PES and EFM-ES perform best among all methods for SCTP dataset in terms of  average MAPE and  average MAE,  respectively,  while the logFM-PES and log-ES dominate all other methods for FF dataset in terms of average MAPE and average MAE, respectively.    

\begin{table}[htbp]
	\centering
	%\begin{small}
	\caption{Forecasting test results of average MAPE and MAE over five runs for SCTP and FF datasets } \label{tab:forectRE}
	\resizebox{0.75\columnwidth}{!}{
		\begin{tabular}{c ccc  ccc }
			\toprule	\midrule
			& \multicolumn{3}{c}{average MAPE} & \multicolumn{3}{c}{ average MAE}  \\
			\cmidrule(lr{1em}){2-4} \cmidrule(rl{1em}){5-7}  			
			& SCTP-P &  SCTP-M  &  FF & SCTP-P &  SCTP-M  &  FF \\
			\midrule			
			\textbf{EFM-PES (proposed)} & \textbf{13.5\%}  & \textbf{11.32\%}  &  35.2\%             & 2.05  & 3.0   & 17.00 \\
			\textbf{EFM-ES (proposed)} &   14.26\%    & 13.51\%                          &  37.5\%              & \textbf{1.02}  & \textbf{1.05}   & 16.25 \\
			logFM-PES                             &  15.38\%    &  15.24\%                         &  \textbf{34.3\%}                & 2.50              & 3.6    & 15.25 \\
			logFM-ES                             &   16.51\%    & 17.5\%                        &  35.5\%              & 2.05             & 3.3     & \textbf{14.25} \\
			SVR                                  &   21\%     & 11.5\%                             &  39.5\%              & 2.7                  & 3.5           & 16.85 \\
			RF                                     & 17.5\%      & 11.5\%                          &  41.5\%                & 1.89                 & 2.05          & 17.25 \\
			\midrule
			\bottomrule
		\end{tabular}%
	} %\end{small}
\end{table}%

\subsection{Discussions}
\subsubsection{Log-transformation v.s.   Exponential  Formulation}
Both log-transformation and exponential formation are  two effective approaches for positive response variables.   Our studies show that %We first note that 
the proposed  exponential formulation methods (EFM-PES and EFM-ES) dominate the log-transformation based methods (logFM-PES and logFM-ES) on sales forecasting and SCTP datasets, however,  for FF dataset,  the log-transformation methods perform better than the exponential formulation based methods.     Intuitively,  log-transformation projects the response variable in another space and  changes the variance.   We doubt that the performance differences might be explained by the distribution of response variable. 
We provide the empirical  distribution of  response variables in Table \ref{tab:edrv};  for each dataset,  the range of response variable values in both training and test was divided into four equal width intervals; the percentage of how many values falling into each interval was computed.   From this table, we observe that  (1) the distribution of Retail Sales are approximately to the uniform distribution; (2)  the Portugueses and Math G3 Grades follow  approximate normal distributions; (3)  the Burned Area in the  FF dataset  is  significantly  different from the others, and highly right-skewed (or positively skewed) distributed.     
Therefore, we suspect that log-transformation is more effective for  datasets with responses following highly  right-skewed distributions.   
%The important insight here is that the log-transformation works when the response variable is highly right-skewed distributed. 
For positive response variables which are not right-skewed distributed,  using exponential formulation as the model is better than using  the log-transformation as the data preprocessing step which might changes the variance of response variables resulting in insufficient training.  
%\blue{
\begin{table}[htbp]
	\centering
		\caption{Empirical distribution of response variables of both training and test in all studied datasets} \label{tab:edrv}%
\resizebox{0.85\columnwidth}{!}{
	\begin{tabular}{r rrrr }
		\toprule
		\multirow{2}{*}{Response Variable and DataSet} & \multicolumn{4}{c}{Interval} \\
		\cline{2-5}
                 & [Min,  0.25Max] & [0.25Max,   0.5Max] & [0.5Max, 0.75Max] & [0.75Max, Max]  \\
         \midrule 
		Retail Sales in 69-y6 & 20.58\% & 29.35\% & 28.53\% & 21.54\% \\
		Retail Sales in p-1 & 28.11\% & 22.59\% & 23.25\% & 26.05\%   \\
		Retail Sales in  x-9w & 23.60\% & 27.52\% & 26.85\% & 22.03\%   \\
		Portuguese  G3 Grade in  SCTP  & 2.46\% & 12.94\% & 64.41\% & 20.19\%  \\
		Math  G3  Grade in SCTP  & 11.64\% & 35.44\% & 42.78\% & 10.12\%   \\
		Burned Area in FF  & \textbf{99.43\%} & \textbf{0.19}\% & \textbf{0.19\%}  & \textbf{0.19\%}   \\
		\bottomrule
	\end{tabular}%
}
\end{table}%

\subsubsection{PES  Loss  v.s. ES Loss} 
Now we investigate the difference between PES and ES minimization for the proposed EFM %and GFM-E 
model on all these datasets.  %estimation on synthetic data sets  to highlight the relationship between the gap and the ratio.  
By results over test sets reported at Table  \ref{tab:forectR} and Table \ref{tab:forectRE},  EFM-PES  (or EFM-ES) performs better than EFM-ES (or EFM-PES) in terms of MAPE  (or MAE) for most cases (except   for 69-y6 where EFM-PES performs better than EFM-ES even in terms of  MAE ).  This results support their effectiveness.      Now we analyze the difference in training process between these  two loss functions.   By Theorem \ref{theorem:es_pes},  the difference is proved to be related to the ratio of maximum response square, $Y^2_{\textrm{max}}$, to minimum response square,  $Y^2_{\textrm{min}}$, of the training set.  (Here we use $Y$ to  denote the general response variable; it refers to sales quantity, G3 grade and burned area for sales, SCTP and FF datasets, respectively).   We have also claimed that PES loss function tends to train model to underestimate data.   Now we provide all related information in Table \ref{tab:ExampTwo}  where  training measurements of mean error squares (MES), mean percentage error squares (MPES) and underestimation ratio are provided.   Note that, 
 the underestimation ratio is defined as the fraction of data points fitted (trained) of which are less than the ground-truth  (actual); for sales forecasting datasets, here training measurements are for data estimation step only (Step 3 in Algorithm 5) and not for the feature selection;   for SCTP and FF datasets,  five runs of training are performed and here average measurements are presented.      From Table \ref{tab:ExampTwo}, we find that  EFM-PES  minimizes the MPES while EFM-ES minimizes the MES which follows the definition;  the underestimation ratio of the EFM-PES increases with the  $\frac{\textrm{Y}^2_{\textrm{max}}}{\textrm{Y}^2_{\textrm{min}}}$ while that of the EFM-ES is around 50\% for all datasets.  This confirms  Theorem \ref{theorem:es_pes} and that PES trains model for underestimation. 
 
 \begin{table}[htbp]
 	%\begin{small}
 	\caption{Training measurements for PES and ES loss functions for  all datasets}  	\label{tab:ExampTwo}
 	\begin{tabular}{clccc}    
 		\toprule
 		\multirow{2}{*}{Dataset} & \multirow{2}{*}{Training measurement} & \multicolumn{2}{c}{Method} & \multirow{2}{*}{$\frac{\textrm{Y}^2_{\textrm{max}}}{\textrm{Y}^2_{\textrm{min}}}$}\\
 		\cline{3-4}
 		& & EFM-PES & EFM-ES \\
 		\hline			
 		\multirow{3}{*}{69-y6} &  mean error squares (MES) & $2.5 \times 10^2$ & $1.1 \times 10^2$ &  \multirow{3}{*}{$4.0 \times 10^2$} \\
 		&  mean percentage error squares (MPES) &  $5.2 \times 10^{-5}$ &  $ 6.6 \times 10^{-5}$ & \\
 		& underestimation ratio & 0.53 & 0.42\\
 		
 		\hline			
 		\multirow{3}{*}{p-1} & mean error squares (MES)  & $ 4.1 \times 10^2$  & $3.8 \times 10^2$ & \multirow{3}{*}{$1.3 \times 10^3$}  \\
 		& mean percentage error squares (MPES) & $8.3 \times 10^{-5}$ &  $9.7 \times 10^{-5}$ \\
 		& underestimation ratio & 0.64 & 0.48\\
 		
 		\hline			
 		\multirow{3}{*}{x-9w} &  mean error squares (MES) & $ 8.5 \times 10^2$  & $7.1 \times 10^2$ & \multirow{3}{*}{$ 2.3 \times 10^3 $} \\
 		&  mean percentage error squares (MPES) & $1.2 \times 10^{-4}$ &  $3.9 \times 10^{-5}$ & \\
 		& underestimation ratio & 0.62 & 0.46  \\
 		
 		\hline			
 		\multirow{3}{*}{SCTP-P} &  average mean error squares (avg. MES) & $ 2.03 \times 10^2$ & $1.03 \times 10^2$ &  \multirow{3}{*}{$3.61 \times 10^4$} \\
 		&  average mean percentage error squares (avg. MPES) & $6.2 \times 10^{-4}$ &  $ 9.8 \times 10^{-4}$ & \\
 		& average underestimation ratio & 0.85 & 0.52\\
 		
 		\hline			
 		\multirow{3}{*}{SCTP-M} &  average mean error squares (avg. MES) & $3.04 \times 10^2$ & $2.02 \times 10^2$ &  \multirow{3}{*}{$4 \times 10^4$} \\
 		&  average  mean percentage error squares (avg. MPES) & $5.7 \times 10^{-4}$ &  $8.5 \times 10^{-4}$ & \\
 		& underestimation ratio & 0.82 & 0.46 \\
 		
 		\hline			
 		\multirow{3}{*}{FF} &  average mean error squares (avg. MES) & $6.5 \times 10^3$ & $4.3 \times 10^3$ &  \multirow{3}{*}{$1.09 \times 10^{9}$} \\
 		&  average mean percentage error squares (avg. MPES) & $ 8.9 \times 10^{-3}$ &  $ 4.5 \times 10^{-2}$ & \\
 		& underestimation ratio & 0.93 & 0.45 \\
 		%\hline
 		\bottomrule
 	\end{tabular}
 	%\end{small}
 \end{table}
%This is explainable:  training errors of instances with large  response values  contributes larger for the ES loss  than that of instances with small response values; on the other hand,  the PES loss increases more heavily with training errors of instances with small response values than instances with large response values.   %In other words,  instances with large response values dominate the ES  training process while PES training process focuses on the instances with small  response values.   This also supports that the GFM-PES likely generates underestimations.   

\begin{figure}[htbp]
	%\begin{scriptsize}
	\begin{center}
		\begin{tabular}{cc}		
			\includegraphics[width=65mm]{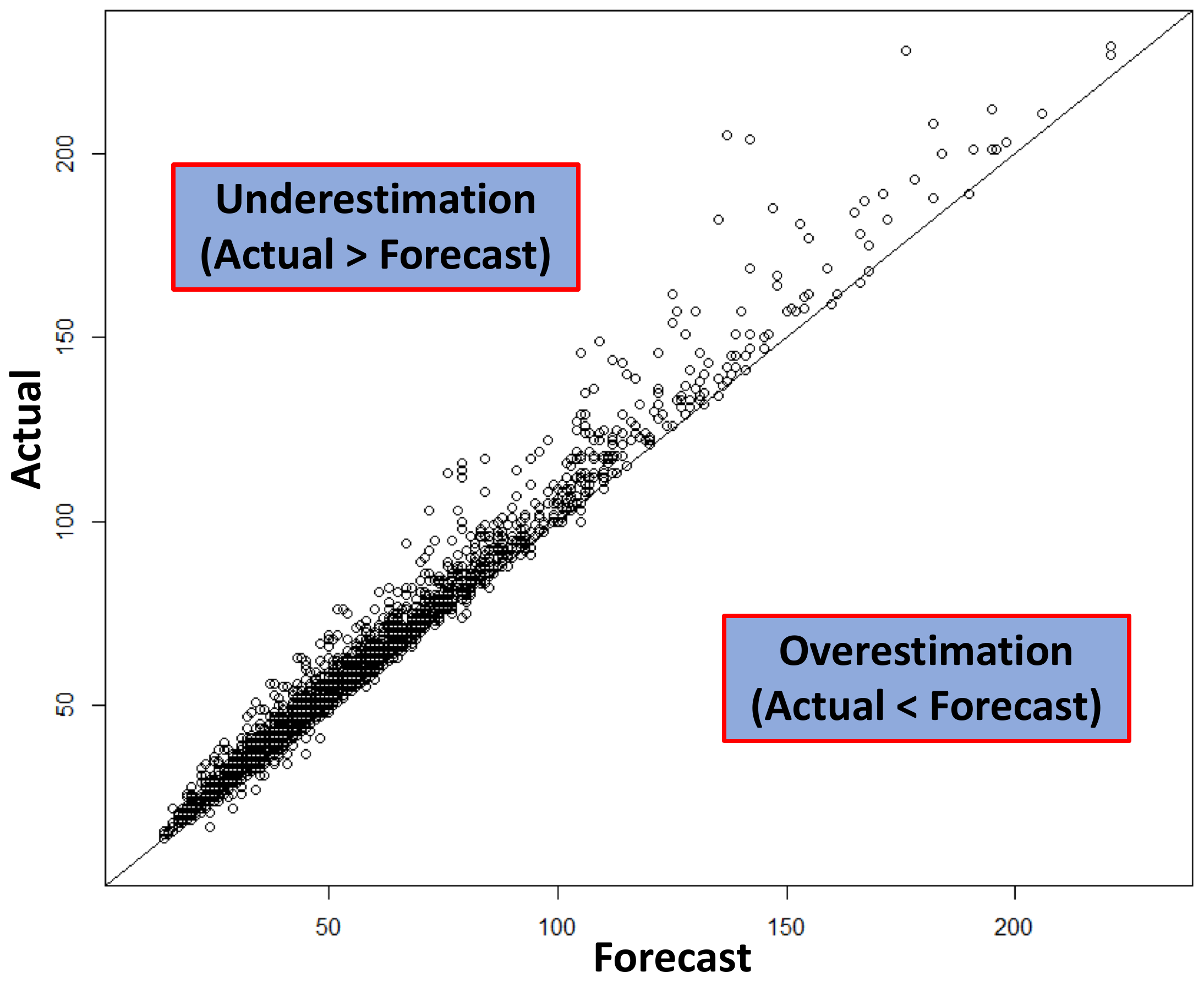}&
			\includegraphics[width=65mm]{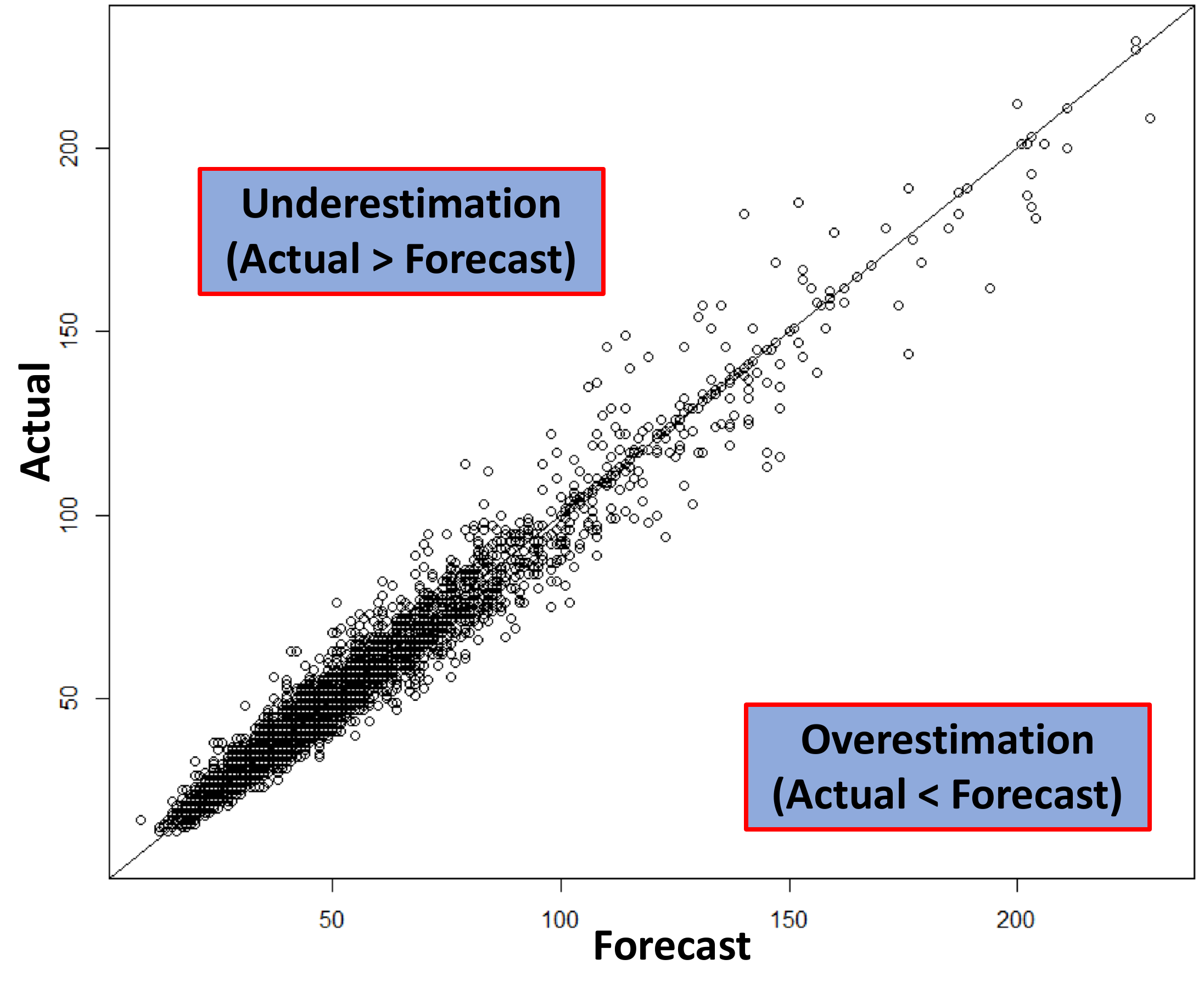}\\
			(a) EFM-PES  &  (b) EFM-ES  \\						
		\end{tabular}		
	\end{center}
	\caption{SKU-store actual sales versus forecast of EFM-PES and EFM-ES for test observations of all three classes}\label{fig:overaResbyPESversusESStore}
	%\end{scriptsize}
\end{figure}

Now, we further provide visualization for SKU-store sales  forecasting test results of the three classes at Figure \ref{fig:overaResbyPESversusESStore}  where,   for each test instance $(i,s) \in E$,  the point $(\hat{d}_{is}, d_{is})$ is plotted in the Forecast-Actual  coordination system and compared with the line Forecast =  Actual;  points lying exactly on the solid line Actual = Forecast  represents a perfect prediction; points over the line mean underestimation (Actual > Forecast); points under the line denote overestimation (Actual < Forecast).    %points under the line denote ovdata while the GFM-ES does not.   %which supports Theorem 1.   
From this Figure,  we again observe that the EFM-PES tends to underestimate data while the EFM-ES does not;  another interesting observation is that, overall, forecasting accuracy of EFM-PES is better than that of EFM-ES for instances with small actual responses $d_{is}$, while,  for instances with large actual responses $d_{is}$,   the EFM-ES performs better.  This is also explainable:  training errors of instances with large  response values  contributes larger for the ES loss  than that of instances with small response values; on the other hand,  the PES loss increases more heavily with training errors of instances with small response values than instances with large response values.   %In other words,  instances with large response values dominate the ES  training process while PES training process focuses on the instances with small  response values.   This also supports that the GFM-PES likely generates underestimations.   

%data while the GFM-ES does not.   
In conclusion, the above analysis  evidences the propositions: (1) the EFM-PES model poses an important property of favouring with underestimation, and (2) PES loss function is dominated by instances with small response values while the ES loss function  focuses on instances with large response values.   
%Now we present qualitative comparison analysis for SKU-store results of the three classes for GFM-PES and GFM-ES by Figure \ref{fig:overaResbyPESversusESStore}  where,   for each test instance $(i,s) \in E$,  the point $(\hat{d}_{is}, d_{is})$ is plotted in the Forecast-Actual  coordination system.% points lying exactly on the solid line Actual = Forecast  represents a perfect prediction; points over the line mean underestimation (Actual > Forecast); points under the line denote overestimation (Actual < Forecast).   
%By Figure \ref{fig:overaResbyPESversusESStore}, we can observe that the GFM-PES tends to underestimate
%data while the GFM-ES does not.   %which supports Theorem 1.   
%Another observation is that forecasting accuracy of GFM-PES is better than that of GFM-ES for instances with small actual responses $d_{is}$, while,  for instances with large actual responses $d_{is}$,   the GFM-ES performs better. % than GFM-PES.    
%This is explainable:  training errors of instances with large $d_{is}$ contributes larger for the ES loss  than that of instances with small $d_{is}$; on the other hand,  the PES loss increases more largely with training errors of instances with small $d_{is}$ than instances with large $d_{is}$.   In other words,  instances with large $d_{is}$ dominate the ES  training process while PES training process focuses on the instances with small $d_{is}$.   This also supports that the GFM-PES likely generates underestimations. 

\subsection{Instance Normalization for PES Minimization of Linear Models } 
In closing this section, we propose a new simple yet effective data normalization for PES minimization for linear models which is motivated by the popular and classic demand-price relationship for pricing and revenue management in supply chain management and marketing area \cite{simchi2005logic}.    %which has wide applciations in supply chain management.    
The linear model  $\mathbf{d} = \mathbf{\beta_0} + \mathbf{\beta_1}\mathbf{x}$ is assumed with  %Settings of $\beta_0$ and $\beta_1$ are subject to 
two  constraints: (1) $\mathbf{d} \geq 0$ and (2) $\mathbf{d}(x)$ is strictly decreasing in $x$, $\beta_1 < 0$.   %It is motivated by the popular demand-price relationship for pricing and revenue management in supply chain management \cite{simchi2005logic}.     
Without loss of generality, we set $\beta_0$ and $\beta_1$ to be 1200 and -10, respectively.   A synthetic data set of 100 points $\{(x_i, d_i) | i = 1, \ldots, 100 \}$ are created:  $x_i$ is sampled uniformly on [1, 50] while $d_i = 1200 - 10 x_i + \epsilon_i$ and noise $\epsilon_i$ follows Gaussian distribution with mean 0 and standard deviation $\sigma$,  $\epsilon_i \sim \mathcal{N}(0, \sigma)$.  The variance of data points is subject to $\sigma$. 

We minimize the ES and PES loss function to estimate parameters of the linear model and study the differences between them.  For the ES minimization, there exists an analytical solution and parameters can be directly derived by least square (LS).  For the PES minimization, note that $\sum^{100}_{i=1} (\frac{\hat{d}_i - d_i}{d_i})^2$ is equivalent to $\sum^{100}_{i=1}  (\frac{\hat{d}_i}{d_i} - 1)^2$.  We propose %a method of 
the least percentage square (LPS) method for parameter estimation for PES minimization and it consists %consisting 
of two steps as follows. %to estimate parameters for PES loss function. 
\begin{enumerate}
	\item Normalization: $\{(x_i, d_i) | i = 1, 2, \ldots, 100\}$ is normalized to $\{(x_i/d_i, 1) | i = 1, 2, \ldots, 100\}$
	\item Apply least square solution on the normalized data set. 
\end{enumerate}

This normalization, ``\textit{instance/sample/row normalization}", unifies the response to be 1 and divides explanatory variables of each data point by its by the  response.  %This technique can be applied whenever percentage (or ratio)  error minimization is the target for parameter estimation under the least square framework.  
It is a new normalization technique and different from popular and classic normalization methods in data mining literature \citep{han2011data} where data are usually normalized per feature/column/variable.

\begin{figure}[htbp]
	%\begin{scriptsize}
	\begin{center}
		\begin{tabular}{cc}
			\includegraphics[width=70mm]{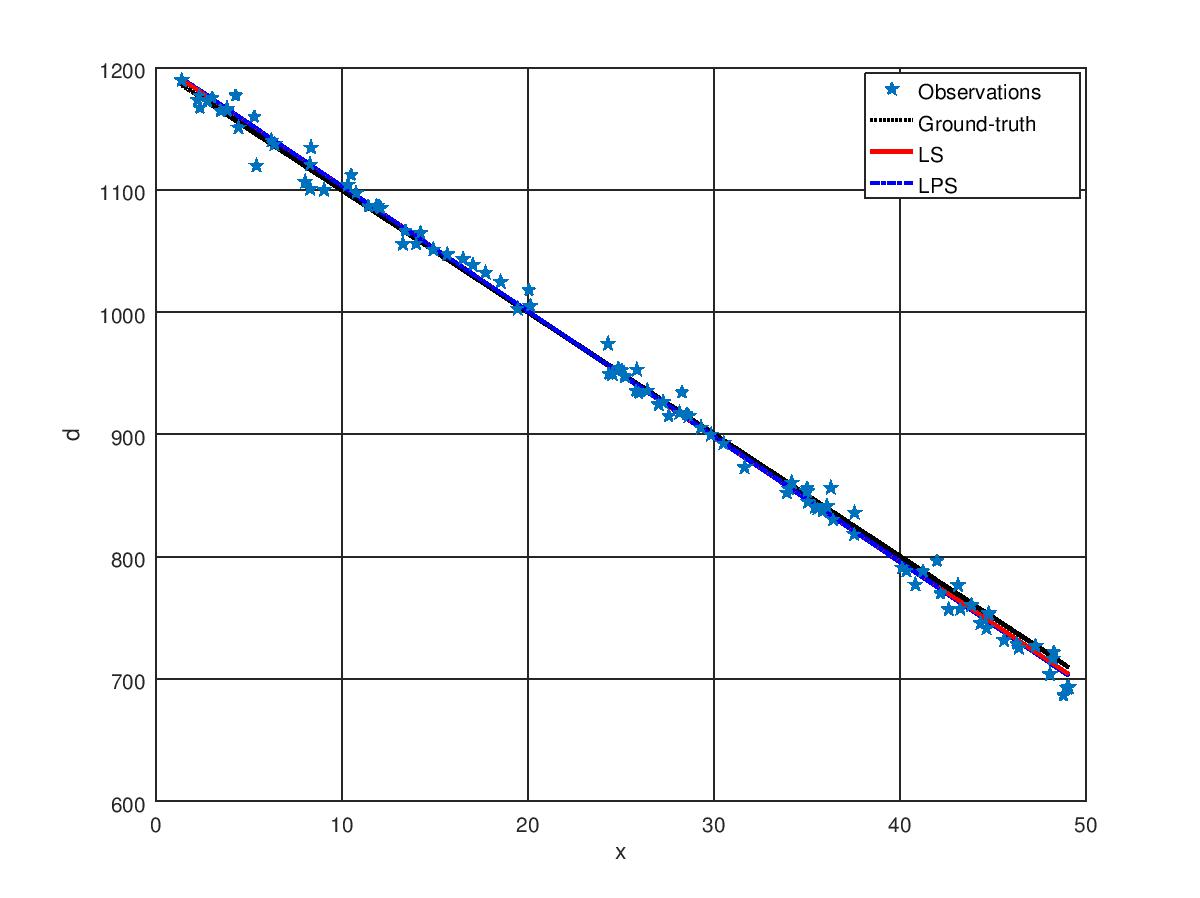}&
			\includegraphics[width=70mm]{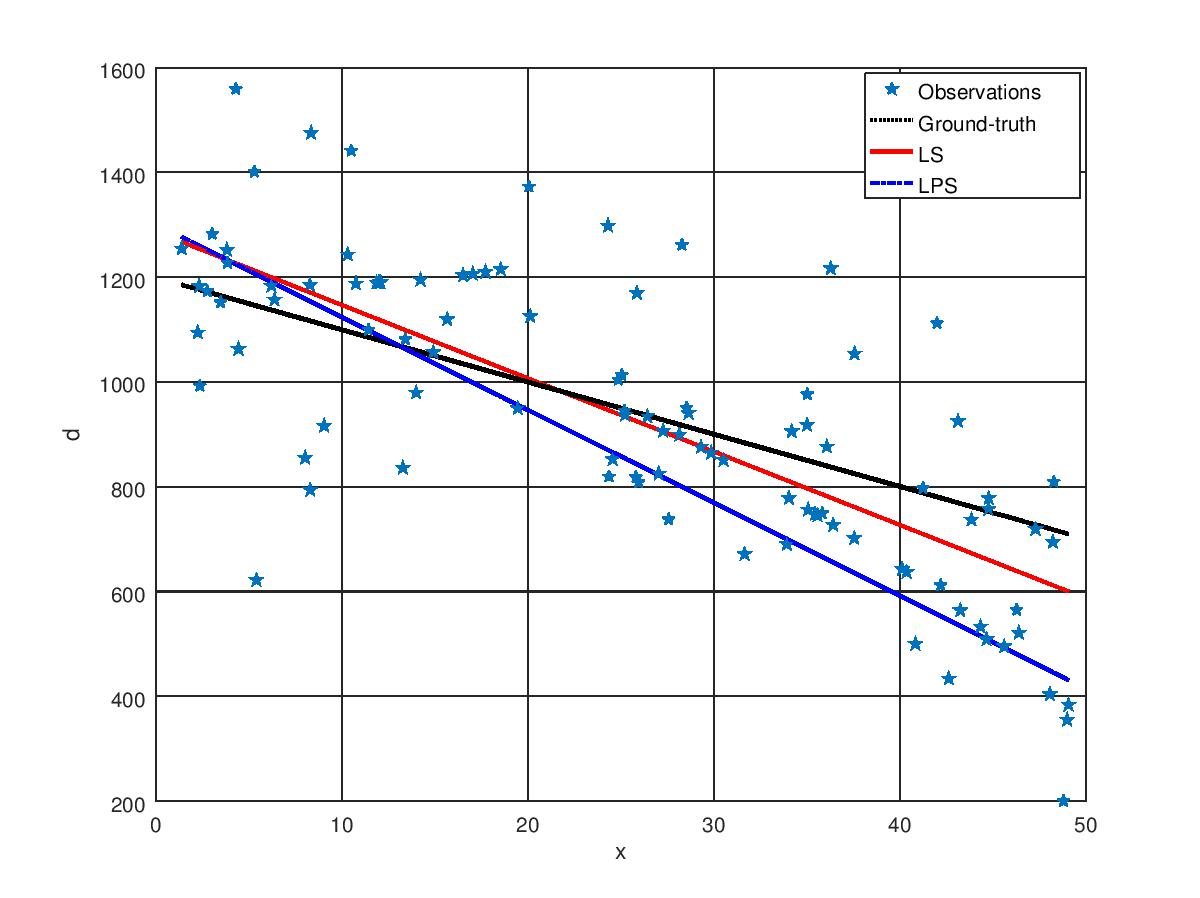}\\
			(a) $\sigma = 10, \frac{d^2_{\textrm{max}}}{d^2_{\textrm{min}}} = 3$ &  (b) $\sigma = 200, \frac{d^2_{\textrm{max}}}{d^2_{\textrm{min}}} = 60.7$  \\						
		\end{tabular}		
	\end{center}
	%\end{scriptsize}	
	\caption{LS and LPS regression on two data sets.}\label{fig:comLSLPS}
\end{figure}

\begin{figure}[htp]
	%\begin{scriptsize}
		\begin{center}
			\begin{tabular}{ccc}
				%				\caption{MSE versus $\frac{d^2_{\textrm{max}}}{d^2_{\textrm{min}}}$}\label{fig:MSEvRatio} &
				%				\caption{MPSE versus $\frac{d^2_{\textrm{max}}}{d^2_{\textrm{min}}}$}\label{fig:MPSEvRatio} \\
				\includegraphics[height = 60mm, width = 45mm]{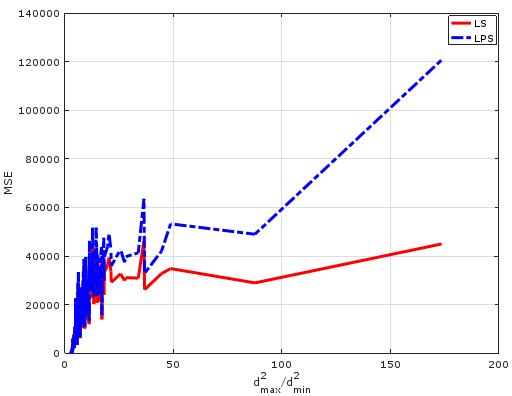} & 
				\includegraphics[height = 60mm, width = 45mm]{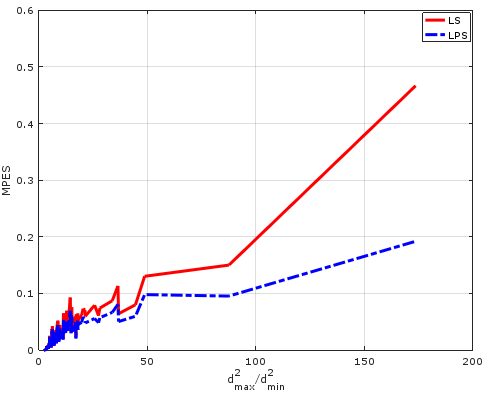} & 		
					\includegraphics[height = 60mm, width = 45mm]{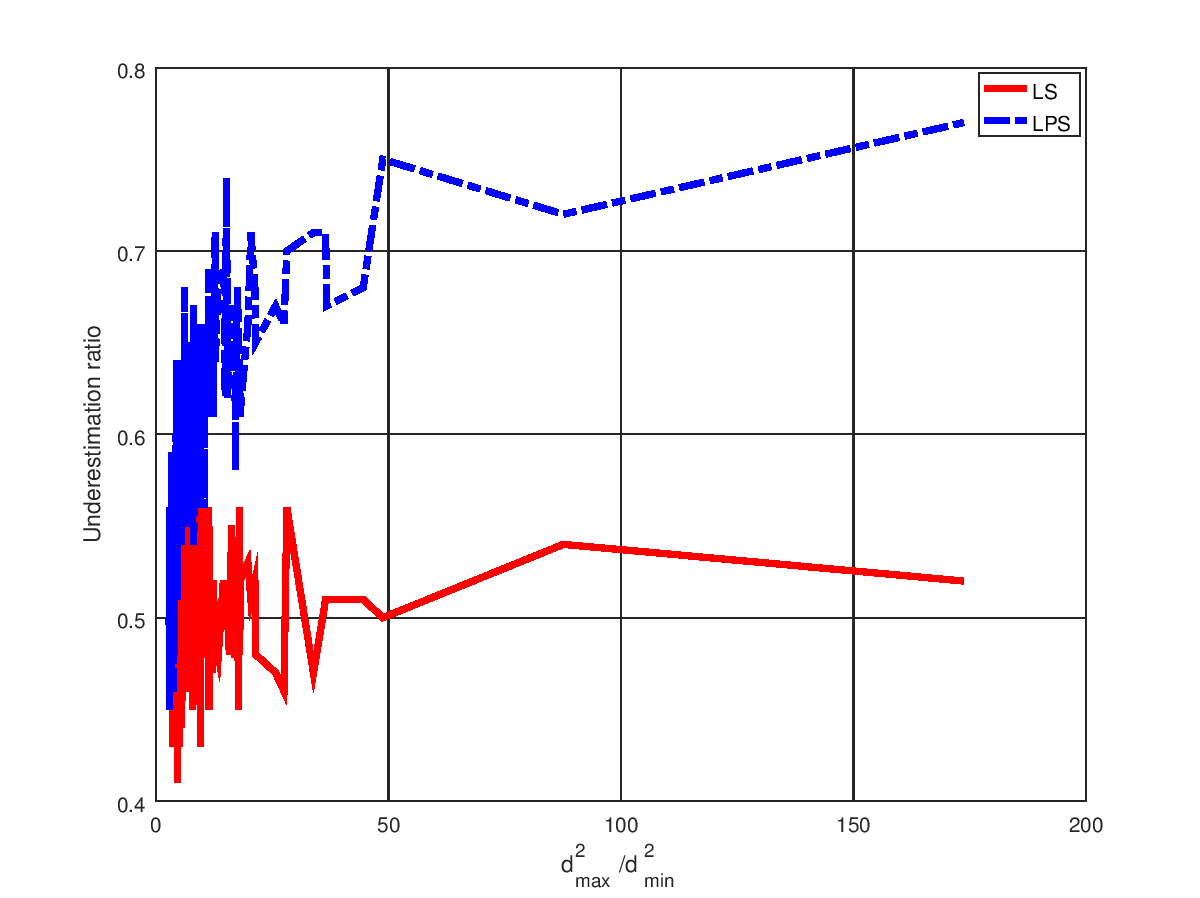} \\			
				(a) MSE versus $\frac{d^2_{\textrm{max}}}{d^2_{\textrm{min}}}$  &  (b) MPSE versus $\frac{d^2_{\textrm{max}}}{d^2_{\textrm{min}}}$ & (c) Underestimation ratio versus$\frac{d^2_{\textrm{max}}}{d^2_{\textrm{min}}}$ \\	
			\end{tabular}		
		\end{center}
	%\end{scriptsize}
	\caption{MSE, MPSE, Underestimation Ratio versus $\frac{d^2_{\textrm{max}}}{d^2_{\textrm{min}}}$} \label{fig:MSEvRatio}	
\end{figure}

Now we demonstrate the effectiveness and differences of the LPS and LS methods on two synthetic data sets of $\sigma = 10$ and $\sigma = 200$.  Figure \ref{fig:comLSLPS} displays the fit results.  
% are tested and fitted results by LS and  LPS are plotted by Figure \ref{fig:comLSLPS}.  Now we investigate the difference between LS and LPS estimation on synthetic data sets  to highlight the relationship between the gap and the ratio.  By Theorem \ref{theorem:es_pes}, the gap is related to the ratio of maximum actual sales square, $d^2_{\textrm{max}}$, to minimum actual sales square,  $d^2_{\textrm{min}}$, and very general but loose bounds are constructed.  Two data sets of $\sigma = 10$ and $\sigma = 200$ are tested first. % for illustration purpose. %to illustrate the idea.  Fit results are plotted in Figure \ref{fig:comLSLPS} and measurements are summarized in Table \ref{tab:ExampTwo}.  
In Figure \ref{fig:comLSLPS} (a), the standard deviation $\sigma$ is 10 and $\frac{d^2_{\textrm{max}}}{d^2_{\textrm{min}}}$ is 3;  no big difference between LS and LPS regression lines are observed.  But if the standard deviation $\sigma$ is increased to 200 and   $\frac{d^2_{\textrm{max}}}{d^2_{\textrm{min}}}$ becomes 60.7 as shown in Figure \ref{fig:comLSLPS} (b), there is a big difference between LPS line which is dotted in blue and the LS line which is solid red; and clearly the blue LPS regression line underestimate the overall data.  
%Further,  the LPS line underestimates most data points as discussed in Section \ref{se:lp:lf}.   Table \ref{tab:ExampTwo} shows fit measurements where the underestimation ratio is defined as the fraction of data points forecast of which are less than the ground-truth  (actual).  For the two data sets,  the LS method produces minimal mean error squares (MES)  while the LPS method generates minimal mean percentage error squares (MPES).   Both MES and MPES gap between LS and LPS are increased with $\frac{d^2_{\textrm{max}}}{d^2_{\textrm{min}}}$.  For data set with $\sigma = 200$, the LPS method underestimates 69\% data points while only 47\% are done by LS method.  
By Theorem \ref{theorem:es_pes}, % the relationship between the derived 
the indicator $\frac{d^2_{\textrm{max}}}{d^2_{\textrm{min}}}$ can effectively  project the  difference between LS and LPS regression.  To investigate it,  
%and underestimation ratio,  fit measurement MES and MPES %gaps of LS and LPS methods,  
we generate 200 data sets with $\sigma$ varying from 1 to 200 with increase of 1, and compute MES,  MPES, underestimation ratio and $\frac{d^2_{\textrm{max}}}{d^2_{\textrm{min}}}$.   %The results are plotted by 
Figure \ref{fig:MSEvRatio}  displays the results.   Obviously, the gap of MES, MPES and underestimation ratio between LS and LPS increases with indicator $\frac{d^2_{\textrm{max}}}{d^2_{\textrm{min}}}$.  %change of MES, MPES, underestimation ratio of LS and LPS with the indicator $\frac{d^2_{\textrm{max}}}{d^2_{\textrm{min}}}$.  
This visualization result demonstrates that  $\frac{d^2_{\textrm{max}}}{d^2_{\textrm{min}}}$ can effectively indicate  the difference between LS and LPS regression for linear models.  
%and \ref{fig:URvRatio}.   We observe that gaps of both MES and MPES between LS and LPS are increased with $\frac{d^2_{\textrm{max}}}{d^2_{\textrm{min}}}$. This indicates that the quantity $\frac{d^2_{\textrm{max}}}{d^2_{\textrm{min}}}$ is a good measure for differences between the LS and LPS method.   Figure \ref{fig:URvRatio} illustrates that the LPS method likely underestimates data points. 

%To be revised due to format issues
%31 March 2019

%\begin{figure}
%	\begin{scriptsize}
%		\begin{center}
			%\begin{tabular}{c}
%			\includegraphics[height = 90mm, width = 100mm]{Figures/Simulation/UnderestimationRatio.png} 			   
			%\end{tabular}	
%		\end{center}
%	\end{scriptsize}
%	\caption{Underestimation ratio versus$\frac{d^2_{\textrm{max}}}{d^2_{\textrm{min}}}$}\label{fig:URvRatio}	
%\end{figure}

\section{Conclusions} \label{sec:con}
In this paper we study the sales forecasting problem for new products with long lead time but short product life cycle. The challenge of this problem is two-fold: (1) long lead time and short product life cycle and (2) no historical sales data for the new items.  An EFM sales forecast model is developed to address it by taking attributes and their interactions into account. 
In order to estimate parameters, we minimize two loss functions of percentage error squares (PES) and error squares (ES) over the training set.   A real-world data set provided by a footwear retailer in Singapore and two public datasets are used to test the proposed approach and the effectiveness has been demonstrated. 
%The forecasting performance in terms of popular techniques, mean absolute percentage error (MAPE) and mean absolute error (AE), are more superior than benchmark regression models and results of existing studies of sales and demand forecasting. 
Several important takeaways from this study are provided below.

The first takeaway is that sales forecasting for new items should take attribute interactions into account.  Moddelling sales by attributes is not a new idea in marketing literature.  Yet, few studies take this into account.  For instance, although Fisher and Vaidyanathan \cite{fisher2014demand} took interactions into account in the implementation, their model somehow ignore them.  Our study considers not only attributes but also their interactions. 

Another takeaway  from this study is the identified differences between PES and ES minimization for parameter estimation.   Minimizing the PES for parameter estimation of regression  models  results  in underestimation which fits the situation where unit-holding cost is much greater  than  unit-shortage cost  (e.g. perishable  products). Moreover, minimizing the PES for linear  regression  models  can  be easily solved by a  ``\textit{instance/sample/data point/row}'' normalization technique  with  classic least  squares  solution  which differs from the popular ``\textit{feature/column/variable}'' normalization.  %Minimizing PES for parameter estimation of regression models results in underestimation which fits the situation where unit-holding cost is much greater than unit-shortage cost (e.g, perishable products).  Moreover, minimizing PES for linear regression models can be easily solved by a novel ``\textit{instance/sample/data point/row}'' normalization technique with classic least square solution which differs from  popular  ``\textit{feature/column/variable}'' normalization.  

The last takeaway is that exponential formulation is better than conventional log-transformation for modeling positive but not skewed distributed response variables, while log-transformation may be employed when the positive responses follow  highly right-skewed distributions.    %And the attribute-level formulation is 

Last but not least, we close this  section  by  pointing   out  avenues  for  future   research.   Integrating replacement behavior into the EFM mode for substitutable products is a potential extension.  This allows replacement behavior considerations to happen not only between attribute levels but also between the interactions.   Another potential extension is the investigation of probabilistic explanation for  the EFM model and theoretically quantifying  the training gap between exponential  formulation and the log-transformation of the distributional skewness of positive response variables. 

\section*{Acknowledgements}
The authors thank three anonymous referees for their constructive comments.   This work is supported by NRF Singapore [Grant NRF-RSS2016-004],  MOE-AcRF-Tier 1 [Grants R-266-000-096-133, R-266-000-096-731, R-266-000-100-646 and R-266-000-119-133], and National Natural Science Foundation of China (Nos. 71801124).

%
% The next two lines define the bibliography style to be used, and the bibliography file.
\bibliographystyle{ACM-Reference-Format}
\bibliography{SalesForecasting}

\appendix
%\section{Proof of Theorem \ref{theorem:es_pes}}  \label{proof:theorem:es:ps}
\section{Proof of Theorem 3.1}  \label{APP:proof:theorem:es:ps}
\proof{We}  start with proving part (a). 
(a) First, by definition of the ES minimizer $\Theta^{*}_{\textrm{ES}}$ in equation (\ref{eq:es:miner}),  
it is true that:
\begin{equation} \label{proof:theorem:0}
	\mathcal{L}^{\textrm{ES}}(\Theta^{*}_{ES}) \leq \mathcal{L}^{\textrm{ES}}(\Theta^{*}_{PES}). 
\end{equation}
Now we prove the second part in (a) which is $\mathcal{L}^{\textrm{ES}}(\Theta^{*}_{PES}) \leq \frac{d^2_{\max}}{d^2_{\min}} ~ \mathcal{L}^{\textrm{ES}}(\Theta^{*}_{ES})$.  By definition of $\Theta^{*}_{\textrm{PES}}$ in equation  (\ref{eq:pes:miner}),  
we have:
\begin{equation} \label{proof:theorem:1}
	\mathcal{L}^{\textrm{PES}} (\Theta^{*}_{\textrm{PES}}) \leq  \mathcal{L}^{\textrm{PES}} (\Theta^{*}_{\textrm{ES}}).
\end{equation}
Extend the left part and we get: 
\begin{equation}  \label{proof:theorem:2}
	\begin{split}
		\mathcal{L}^{\textrm{PES}} (\Theta^{*}_{\textrm{PES}}) ~ & = \sum_{(i, s) \in T} {[\frac{\hat{d}_{is} - d_{is}}{d_{is}}]^2} \\
		&  = \sum_{(i, s) \in T} {\frac{[f(\mathbf{x}^{is}; \Theta^{*}_{\textrm{PES}}) - d_{is}]^2}{d^2_{is}}} \\
		& \geq  \sum_{(i, s) \in T} \frac{[f(\mathbf{x}^{is}; \Theta^{*}_{\textrm{PES}}) - d_{is}]^2}{d^2_{\max}} \\
		& =  \frac{1}{d^2_{\max}}  \sum_{(i, s) \in T} [f(\mathbf{x}^{is}; \Theta^{*}_{\textrm{PES}}) - d_{is}]^2 \\
		& =  \frac{1}{d^2_{\max}} \mathcal{L}^{\textrm{ES}}(\Theta^{*}_{\textrm{PES}}),
	\end{split}
\end{equation}
where the inequality follows from the fact that, for each $(i, s) \in T$,  $d_{\max} \geq d_{is} \geq 1 $, $\frac{1}{d^2_{is}} \geq \frac{1}{d^2_{\max}} > 0$ and $[f(\mathbf{x}^{is}; \Theta^{*}_{\textrm{PES}} - d_{is})]^2 \geq 0$; and the first and the last equivalences are from  definitions of $l^{\textrm{PES}}(\cdot)$ and  $l^{\textrm{ES}}(\cdot)$.

Extend the right part of equation (\ref{proof:theorem:0}) and we obtain: 
\begin{equation}  \label{proof:theorem:3}
	\begin{split}
		\mathcal{L}^{\textrm{PES}} (\Theta^{*}_{\textrm{ES}}) ~ & = \sum_{(i, s) \in T} {[\frac{\hat{d}_{is} - d_{is}}{d_{is}}]^2} \\
		&  = \sum_{(i, s) \in T} {\frac{[f(\mathbf{x}^{is}; \Theta^{*}_{\textrm{ES}}) - d_{is}]^2}{d^2_{is}}} \\
		&  \leq \sum_{(i, s) \in T} {\frac{[f(\mathbf{x}^{is}; \Theta^{*}_{\textrm{ES}}) - d_{is}]^2}{d^2_{\min}}} \\
		& =  \frac{1}{d^2_{\min}}  \sum_{(i, s) \in T} [f(\mathbf{x}^{is}; \Theta^{*}_{\textrm{ES}}) - d_{is}]^2 \\
		& =  \frac{1}{d^2_{\min}} \mathcal{L}^{\textrm{ES}}(\Theta^{*}_{\textrm{ES}}),
	\end{split}
\end{equation}
where, similarly,  the inequality follows from the fact that, for each $(i, s) \in T$,  $d_{is} \geq d_{\min} \geq 1 $, $\frac{1}{d^2_{\min}} \geq \frac{1}{d^2_{is}} > 0$ and $[f(\mathbf{x}^{is}; \Theta^{*}_{\textrm{ES}} - d_{is})]^2 \geq 0$; and the first and the last equivalences are from definitions of $l^{\textrm{PES}}(\cdot)$ and  $l^{\textrm{ES}}(\cdot)$. % given by equations (\ref{eq:error:pes}) and (\ref{eq:error:es}), respectively.  

By equations  (\ref{proof:theorem:1}), (\ref{proof:theorem:2}) and (\ref{proof:theorem:3}), we get: 
\begin{equation}
	\frac{1}{d^2_{\max}}  \mathcal{L}^{\textrm{ES}}(\Theta^{*}_{\textrm{PES}}) \leq \mathcal{L}^{\textrm{PES}} (\Theta^{*}_{\textrm{PES}}) \leq \mathcal{L}^{\textrm{PES}} (\Theta^{*}_{\textrm{ES}}) \leq  \frac{1}{d^2_{\min}} \mathcal{L}^{\textrm{ES}}(\Theta^{*}_{\textrm{ES}}), 
\end{equation}
which is simplified by $d^2_{\max} > 0$: 
\begin{equation}
	\mathcal{L}^{\textrm{ES}}(\Theta^{*}_{\textrm{PES}}) \leq  \frac{d^2_{\max}}{d^2_{\min}} \mathcal{L}^{\textrm{ES}}(\Theta^{*}_{\textrm{ES}}). 
\end{equation}
Therefore, part (a) in Theorem \ref{theorem:es_pes} is proved. 

Now we prove part (b). First, by equation (\ref{eq:pes:miner}),  because $\Theta^{*}_{\textrm{PES}}$ is the minimizer of $\mathcal{L}^{\textrm{PES}}(\Theta)$, 
\begin{equation}
	\mathcal{L}^{\textrm{PES}}(\Theta^{*}_{PES}) \leq \mathcal{L}^{\textrm{PES}}(\Theta^{*}_{ES}). 
\end{equation}
Now we recall equation (\ref{proof:theorem:0}), that is: 
\begin{equation} \label{theorem:1:b-1}
	\sum_{(i, s) \in T} [f(\mathbf{x}^{is}; \Theta^{*}_{\textrm{ES}}) - d_{is}]^2 \leq \sum_{(i, s) \in T} [f(\mathbf{x}^{is}; \Theta^{*}_{\textrm{PES}}) - d_{is}]^2.
\end{equation}
Dividing both sides of the inequality by positive value $\frac{1}{d^2_{\min}}$:
\begin{equation} \label{theorem:1:b-2}
	\frac{1}{d^2_{\min}} \sum_{(i, s) \in T} [f(\mathbf{x}^{is}; \Theta^{*}_{\textrm{ES}}) - d_{is}]^2  
	\leq 
	\frac{1}{d^2_{\min}} \sum_{(i, s) \in T} [f(\mathbf{x}^{is}; \Theta^{*}_{\textrm{PES}}) - d_{is}]^2.
\end{equation}
Now we extend left side of the inequality:
\begin{equation}  \label{theorem:1:b-3}
	\begin{split}
		\frac{1}{d^2_{\min}} \sum_{(i, s) \in T} [f(\mathbf{x}^{is}; \Theta^{*}_{\textrm{ES}}) - d_{is}]^2 ~ 
		& = \sum_{(i, s) \in T} \frac{[f(\mathbf{x}^{is}; \Theta^{*}_{\textrm{ES}}) - d_{is}]^2}{d^2_{\min}}  \\
		& \geq \sum_{(i, s) \in T} \frac{[f(\mathbf{x}^{is}; \Theta^{*}_{\textrm{ES}}) - d_{is}]^2}{d^2_{is}}\\
		& =  \mathcal{L}^{\textrm{PES}}(\Theta^{*}_{\textrm{ES}}),
	\end{split}
\end{equation}
where the first inequality follows from that the domain of $d_{is}$ is $\mathbb{N}_{>0}$; for each  $d_{is}$, $(i, s) \in T$,   $\frac{1}{d^2_{\min}} \geq \frac{1}{d^2_{is}} > 0$ and $[f(\mathbf{x}^{is}; \Theta^{*}_\textrm{ES}) - d_{is}]^2 \geq 0$. The last equality comes from  definition of the PES loss function $\mathcal{L}^{\textrm{PES}}(\cdot)$.

Now we extend the right side of inequality (\ref{theorem:1:b-2}): 
\begin{equation} \label{theorem:1:b-4}
	\begin{split}
		\frac{1}{d^2_{\min}} \sum_{(i, s) \in T} [f(\mathbf{x}^{is}; \Theta^{*}_{\textrm{PES}}) - d_{is}]^2 ~
		& = \frac{d^2_{\max}}{d^2_{\min}} \frac{1}{d^2_{\max}} \sum_{(i, s) \in T} [f(\mathbf{x}^{is}; \Theta^{*}_{\textrm{PES}}) - d_{is}]^2 \\
		& = \frac{d^2_{\max}}{d^2_{\min}} \sum_{(i, s) \in T} \frac{[f(\mathbf{x}^{is}; \Theta^{*}_{\textrm{PES}}) - d_{is}]^2}{d^2_{\max}} \\
		& \leq  \frac{d^2_{\max}}{d^2_{\min}}  \sum_{(i, s) \in T} \frac{[f(\mathbf{x}^{is}; \Theta^{*}_{\textrm{PES}}) - d_{is}]^2}{d^2_{is}} \\
		& =  \frac{d^2_{\max}}{d^2_{\min}} l^{\textrm{PES}}(\Theta^{*}_{\textrm{PES}}),
	\end{split}
\end{equation}
where the inequality is due to that, for $(i, s) \in T$,  $d_{\max} \geq d_{is} \geq 1$,  $\frac{1}{d^2_{is}} \geq \frac{1}{d^2_{\max}} > 0 $ and $[f(\mathbf{x}^{is}; \Theta^{*}_\textrm{ES}) - d_{is}]^2 \geq 0$ 

By inequalities (\ref{theorem:1:b-2}), (\ref{theorem:1:b-3}) and (\ref{theorem:1:b-4}), we can get:
\begin{equation}
	\mathcal{L}^{\textrm{PES}}(\Theta^{*}_{\textrm{ES}}) \leq \frac{d^2_{\max}}{d^2_{\min}} \mathcal{L}^{\textrm{PES}}(\Theta^{*}_{\textrm{PES}}),
\end{equation}
which completes the proof for part (b). 
\endproof

\end{document}